# *Building a Classification Model for Enrollment In Higher Educational Courses using Data Mining Techniques*

**Priyanka Saini**
**BANASTHALI VIDYAPITH, RAJASTHAN**

# Abstract

## *Classification model for the Student's Enrollment process in higher educational courses using data mining techniques*

Data Mining is the process of extracting useful patterns from the huge amount of database and many data mining techniques are used for mining these patterns. Recently, one of the remarkable facts in higher educational institute is the rapid growth data and this educational data is expanding quickly without any advantage to the educational management. The main aim of the management is to refine the education standard; therefore by applying the various data mining techniques on this data one can get valuable information.

This research study proposed the "classification model for the student's enrollment process in higher educational courses using data mining techniques". Additionally, this study contributes to finding some patterns that are meaningful to management.

# ***Table of Contents***

# List of Figures

# *List of Tables*

# CHAPTER 1

INTRODUCTION

## 1.1 Data Mining

### *History of Data Mining*

Data mining was presented in the 1990s.Data mining roots are traced along three family lines: Statistics, Artificial Intelligence (AI), and Machine Learning.

Statistics founds many technologies such as regression analysis, standard distribution, standard variance, standard deviation, cluster analysis, discriminate analysis. The data mining is built on these technologies and these technologies are helpful to study data and data relationships also. Data mining classification algorithms use statistical method to create decision tree and rules to validate machine learning based model.

Artificial Intelligence is based on heuristics and it is used to apply human thought processing to statistical issues. Data Mining adopts ideas from artificial intelligence and pattern recognition fields such as signal processing or visualization techniques.

Machine Learning is combination of Statistics and AI. It is a evolution of artificial intelligence, because it mixes Artificial intelligence heuristics with advanced statistical analysis. Figure 1.1 shows intersection of data mining with various related technologies.

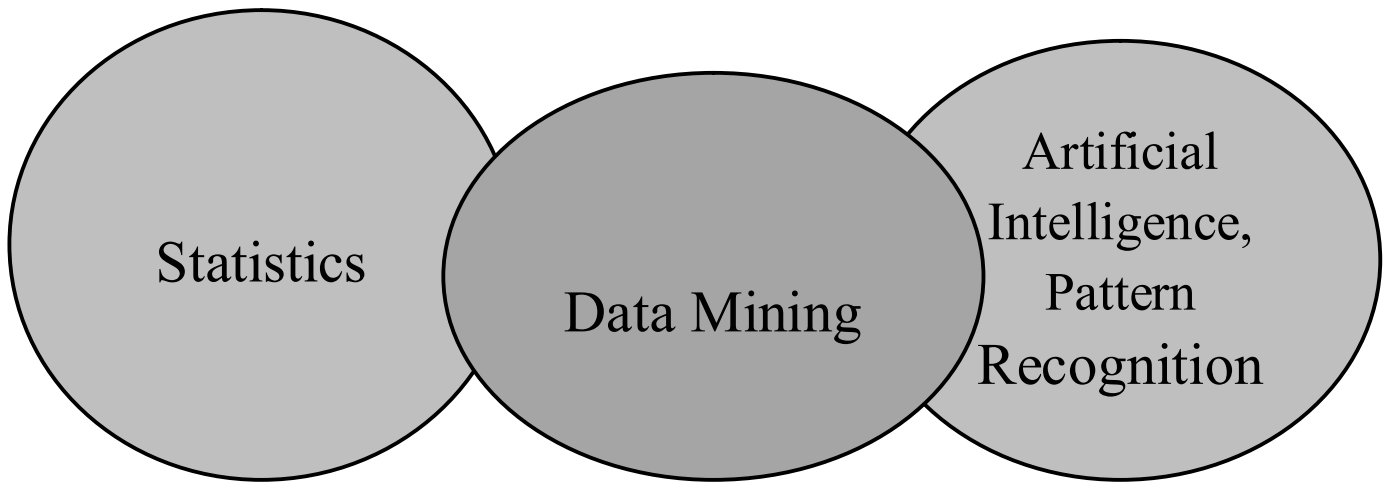

**Figure 1.1: Intersection of technologies yielding to data mining**

### *Data Mining Definition*

A knowledge discovery process in databases consists of these steps:

- Understanding the requirements
- Selecting a target data set
- Integrating the data set
- Data cleaning and preprocessing
- Model building
- Select data mining algorithms
- Evaluating the results
- Testing the results
- Using this discovered knowledge

## *Working of Data mining*

Data mining is the process of analyzing and summarizing data into useful information. Data mining software is an analytical tool that allows analyzing, categorizing and summarizing the data. Data mining process has five elements that can be summarized as:

- **Extraction, transformation, and loading transaction** data onto the data warehouse system.

- **Storing and managing** the data in a database system(multidimensional).

- **Providing data access facility** to business analysts and professionals.

- **Analyzing** the data (by application software)

- **Present** the data in a graph or table.

## *Data Mining Techniques*

- ***Classification:*** Classification is the best & popular approach. Classification is the process of finding a set of models or functions that describe and distinguish data classes or concepts, for the purpose of being able to use the model to predict the class of objects whose class label is unknown, Unlike a classification model, the purpose of prediction model is to determine the future outcome rather than the current behaviour. Its output can be categorical or numeric value. It is a supervised learning because the classes are determined before examining the data.

- ***Clustering:*** Clustering is best suitable for finding groups of similar data items. In Clustering, a set of data items is partitioned into a set of classes such that similar characteristics items are grouped together.

- ***Association analysis:*** It is used to discover relationships between attributes and items such as the presence of one pattern implies the presence of another pattern. Association Rule is a popular technique for market basket analysis because all possible combinations of interesting product groupings can be explored .

- ***Decision Tree:*** It is a tree structure like a flow-chart, in which the rectangular boxes are called the node. Each node represents a set of records from the original data set. Internal node is a node that has a child and leaf (terminal) node is nodes that don't have children. Root node is a topmost node. The decision tree is used for finding the best way to distinguish a class from another class. There are five mostly & commonly used algorithms for decision tree: - *ID3, CART, CHAID, C4.5 algorithm and J48.*

### *Data Mining Advantages*

Data mining is so desired because of its many benefits. We have many cheaper techniques to collect and manage the data, but they are a few techniques for extracting useful knowledge from this data. Data mining has various advantages in many fields.

***Marketing and Retailing:*** Marketers can make policy to fulfill the customer needs and understand their purchasing behaviors with the help of data mining.

***Banking:*** Financial institutions can get help of data mining in credit and loan information. A credit card issuer can detect fraud credit card transaction.

***Researchers:*** Using data mining techniques researchers extract the useful knowledge by analyzing their data and precede their research work.

***Education***: Data mining is very useful in educational institutes because there is a large unused collected data and this data can be used in a proper way using data mining.

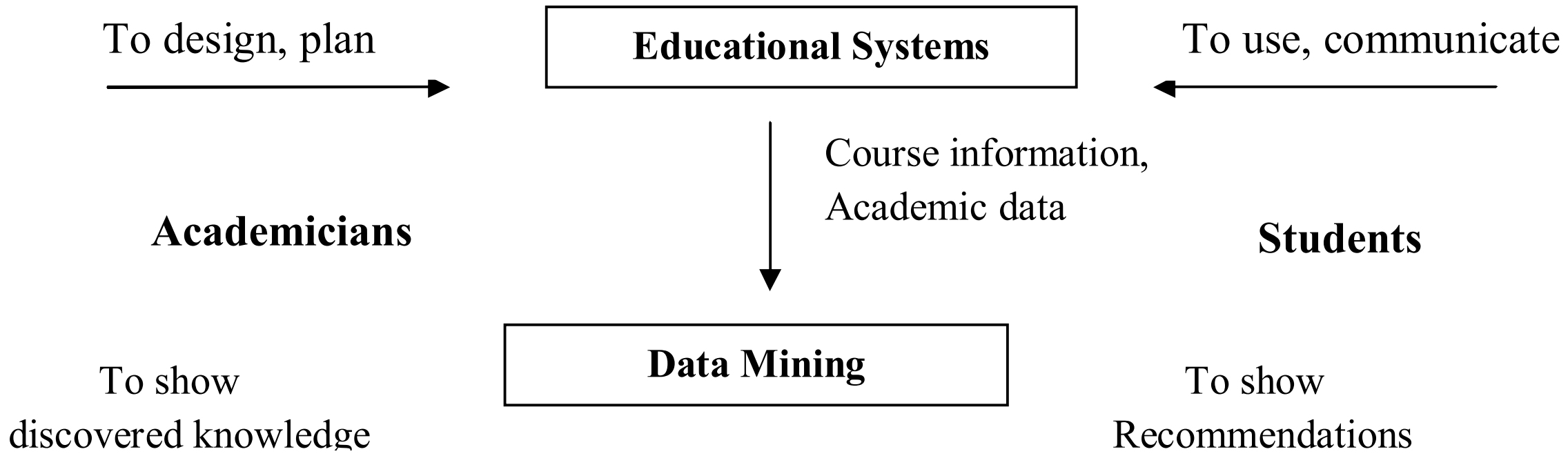

**Figure 1.2 : Data Mining in Education**

# 1.2 Need of Current Research

The educational data contains not only grades, exams or enrollment data but also other information. Attributes obtained from this data analysis are considered to be valuable.

### *Educational Data Mining*

Educational data mining (EDM) combines data mining and knowledge discovery methods into educational setting.EDM explores the row data from educational environment to useful data that can be used to making decisions and solve the problems.

### *A. Data Mining Techniques in Educational Systems*

In the educational systems, data mining techniques can search useful patterns and these useful patterns can be used for many purpose.

***Predicting Student Performance:***
In Educational Systems, using data mining techniques student performance, score values are predicted and these values can be numerical or categorical. Regression analysis finds relationship between a dependent (numerical) and one or more independent variable (numerical).Classification technique is used to classify individual items. Different data mining techniques such as neural network, rule based system, regression analysis, correlation analysis are applied on educational data.

***Grouping Students:***
Clustering and Classification data mining techniques are used to build groups of students based on their characteristics, performance, etc. Different clustering algorithms are K-mean, hierarchical, model-based clustering.

***Enrollment Management:***
Enrollment management is required in educational system to shape the enrollment strategy of an institution and fulfill all goals. It is a set of activities such as marketing, retention program and admission process.

## 1.3 Research Goal

The goal of this thesis can be split as following.

- ***Concept:*** Understand the concept of data analysis and how these methods can be applied on educational data such as classification analysis.

- ***Method:*** After the initial research, select a data mining classification method .

- ***Evaluate:*** Find a method that are able to correctly classify a student data set of with a success rate of at least 70% . For this requirement decision tree method is used here for building a classification model that provide 69.69% model accuracy

## 1.4 Research Contribution

This project aims at developing a classification model for Enrollment in Higher Education Courses using data mining techniques. This model helps us to fulfill these objectives:

### *Classification Analysis on student data*

Helpful to predict student performance from past academic performance, by which-

1) We can identify suitable student stream in a particular course and can make a good proper counseling strategy for a particular course.

2) It can also suggest that which course is applicable for student according to his skills.

### *Advertisement Management*

It is useful to discover some meaningful pattern for institute such as select a best advertisement medium using some data mining methods.

### *Classification of Student's data Using Data Mining Techniques for Training & Placement Purpose.*

## 1.5 Research Motivation

In India, there is largest no. of educational institutes, so it is second largest in the world after United States. There is more competition between all institutes for attracting students to get enrollment in their institutes so they focus on strength of students not quality of education at the time of enrollment.

Today Admission process of institutes has become very critical. There are many problems at the time of admission in institutes because many students apply for courses but seats are limited, so there is no proper seat allocation of courses to the students so students are unable to get enroll in their interested courses. Some students have good marks but they get admission in other course (that is not according to their subjects) due to limited seats.

So there is a proper attention is needed in admission process. Every year huge amount of student data is recorded in database however this data is not put in proper form. There is a requirement of data mining that handle these challenges & overcome them. Then there is enough information for better planning, evaluation and decision making. Data mining will extract hidden information from student enrollment database, this information will be meaningful for institutes.Then a better & mined knowledge is present in database that can be use directly, there is no extra requirement.

The motive behind in this paper is based on classification model for enrollment in higher educational courses using data mining techniques. This is useful for predicting the students that are interested to take admission in higher study course, this is also useful to select a good student for a particular course & also useful to give guidelines to student about selection of course. By this study we will find some meaningful pattern that can be useful for institutes.

## 1.6 Thesis Organization

The organization of the rest of this thesis is as follows:

- Chapter 2 presents the related work
- Chapter 3 discusses the research methodology that presents results, conclusion and future directions about the work done in the thesis
- Chapter4 discusses the References.

# CHAPTER 2

## RELATED WORK

Data mining extract hidden patterns from the student database that provides help to university in education. The educational data mining is a recent and popular area, there are many journals, ongoing books, workshops [1].

**Delavari and Beikzadeh [2]** give knowledge to use data mining methods in Higher learning institutions and define how data mining can be applied to the educational data.

**R. R. Kabra and R. S. Bichkar [3]** define that a model can be created using student's past-academic performance with the help of decision tree algorithm and this model can predict student's performance in the first year of engineering exam.

**Oladipupo and oyelade [4]** show their study in which student's failure patterns are identified using association rule data mining technique. Their study trims down failure rate and improves academic performance.

**Zlatko J. Kovacic [5]** show a case study for student's success prediction using educational data mining. For classify successful and unsuccessful students the *CHAID and CART algorithm* were applied on enrollment data of open polytechnic (New Zealand).

**Nguyen et al. [6]** predict the performance of undergraduate and postgraduate students at two different institutes using decision tree and Bayesian network and this is used to find and helping failing students and determines scholarship. In this result, decision tree gives better accuracy than Bayesian network.

**Hijazi and Naqvi [7]** show a case study on student performance. In this study 300 student's sample is taken from colleges of **Punjab university(Pakistan) and** they determine that high correlation is present between some factor (mother's education factor and student's family income factor) and student performance.

**T.Miranda Lakshmi, A.Martin, R.Mumtaj Begum and Dr.V.Prasanna Venkatesan [8]** conduct a case study on student's qualitative data using decision tree algorithms to identify the effect of qualitative data in the performance of the student.

**Mohammed M. Abu Tair, Alaa M. El-Halees [9]** conduct a case study on graduate students' data using data mining techniques to improve performance and extract useful knowledge from this data.

**Sunita B.Aher, L.M.R.J. lobo [10]** conduct a comparative study to predict course selection using association rule algorithms.

**Umesh Kumar Pandey and S. Pal [11]** administrate a study in which 600 student's data sample is taken from Dr. M.L. Awadh University's colleges ,Faizabad and found that whether new student will perform or not.

**Bhise R. B, Thorat S.S and Supekar A.K. [12]** performed data mining process on the student's database using clustering- K-mean algorithm. **K.Shanmuga Priya and A.V.Senthil Kumar [13]** use a classification method that helps to improve the performance & extract the knowledge from student's final semester marks.

**Varun Kumar and Anupama Chadha [14]** used association rule technique to improve the performance of postgraduate students. They focus on many factors like student's interest, teaching methodologies, curriculum design using association rule mining and these factors can affect post graduate student's performance.

**Abeer Badr El Din Ahmed and Ibrahim Sayed Elaraby [15]** use decision tree technique for data classification that is helpful for predicting the student's final grade.

**In 1986 J.R Quinlan** summarizes an approach and describes ID3 and this was the first research work on ID3 algorithm [16].**Anand Bahety** implemented the ID3 algorithm on the "Play Tennis" database and classified whether the weather is suitable for playing tennis or not?.Their results concluded that ID3 doesn't works well in continuous attributes but gives good results for missing values [17].

**Mary Slocum** gives the implementation of the ID3 in the medical area. She transforms the data into information to make a decision and performed the task of collecting and cleaning the data .Entropy and Information Gain concepts are used in this study [18].

**Kumar Ashok (et.al)** performed the id3 algorithm classification on the "census 2011 of India" data to improving or implementing a policy for right people. The concept of information theory is used here. In the decision tree a property on the basis of calculation is selected as the root of the tree and this process's steps are repeated [19].

**Sonika Tiwari** used the ID3 algorithm for detecting Network Anomalies with horizontal portioning based decision tree and applies different clustering algorithms. She checks the network anomalies from the decision tree then she discovers the comparative analysis of different clustering algorithms and existing id3 decision tree [20].

**Yadav, Bharadwaj and Pal [21]** obtained the students data such as attendance, seminar, assignment marks and class test to predict the end semester performance using three algorithms ID3decision tree, C4.5 and CART and result shows that CART gives better result for classification of data.

**Pandey and Pal [22]** show their study using association rule analysis to find the student interest of choosing class language. In this paper they use seven different interestingness measures. Their result concluded that student has shown their interest in mix mode class language.

**Bharadwaj and Pal [23]** use the classification decision tree technique to evaluate student' end semester performance, this study helps to identify the dropouts and students who require special attention and teacher advising.

**AI-Radaideh. et al [24]** presents a classification based model for student performance prediction using ID3 algorithm,C4.5 and Naïve Bayes algorithm but decision tree had better results.

**K.S. Priya and A.V.S. kumar [25]** use a classification approach that extracts the knowledge from student end semester marks.

Data Mining can be used in educational field to enhance our understanding of learning process to focus on extracting and identifying the variables of the learning process of students as described by **Alaa el- Halees [26].**

**Han and Kamber [27]** describes data mining software that allow the users to analyze data from different views, and summarize these relationships which are identified during the mining process.

**Divakar, R.C Jain [28]** applied four classification methods on student academic data i.e Decision tree (ID3), Multilayers perceptron, Decision table & Naïve Bayes classification method.

**Shaeela Ayesha, Tasleem Mustafa, Ahsan Raza Sattar, and M. Inayat Khan [29]** applied K-mean clustering to analyze learning behavior of students which will help the tutor to improve the performance of students and reduce the dropout ratio to a significant level.**D'Mello [30]** studied on bored and frustrated student.

# CHAPTER 3

## RESEARCH METHODOLOGY

## 3.1 CLASSIFICATION FOR MCA STUDENT DATA

In recent years, Indian higher educational institutes grow rapidly. There is more competition between institutes for attracting students to get enrollment in their institutes. The admission process is conducted every year at the institute and it results in the recording of large amounts of data. But, in most of the cases this data is not properly utilized (or analyzed) and results in wastage of what would otherwise be one of the most precious assets of the institutes. By applying the various data mining techniques on this data one can get valuable information and predictions can be done for the betterment of the admission process. This study presents data mining techniques for the enrollment process in MCA stream. These methods will help to improve the overall performance of the admission process at higher educational institutes.

### 3.1.1 Data Used (Data Collection & Data Preprocessing)

- The MCA student data include: Category, Father Qualification, Mother Qualification, XII Medium, XII Stream, Mathematics Grade in XII, XII Grade, UG Stream, UG Medium, UG Grade, PG Grade.

- Some attributes are selected using attribute selection measure, these attributes are Mathematics Grade in XII, XII Grade, UG Stream, UG Grade, PG Grade.

- The next step of our application focuses on transforming these data in order to be used in Weka, a data mining specialized software. Since ID3 algorithm can only work with categorical variables, we have to adapt the data. With this intention we will make some transformations:

- The Mathematics Grade in XII values will be encoded like: A– (80-89), B – (70-79), C – (60-69), D – (50-59), E – (40-49), F<=35, not applicable

- The XII Grade values will be normalized, after normalization each ones having attached a Grade A– (80-89), B – (70-79), C – (60-69), D – (50-59), E – (40-49)

  After Normalization

  if XII Per <=0 THEN E<br>
  if XII Per <=0.25 THEN D<br>
  if XII Per <=0.487 THEN C<br>
  if XII Per <=0.75 THEN B<br>
  Else A

- The UG Grade values will be normalized, after normalization each ones having attached a Grade A– (80-89), B – (70-79), C – (60-69), D – (50-59), E – (40-49)
  After Normalization

  if UG Per <=0 THEN E
  if UG Per <=0.11 THEN D
  if UG Per <=0.41 THEN C
  if UG Per <=0.70 THEN B
  Else A

- The PG Grade values will be normalized, after normalization each ones having attached a Grade A– (80-89), B – (70-79), C – (60-69), D – (50-59), E – (40-49)
  After Normalization

  if PG Per <=0 THEN E
  if PG Per <=0.23 THEN D
  if PG Per <=0.46 THEN C
  if PG Per <=0.74 THEN B
  Else A

## 3.1.2 Used Tool

### *WEKA:-*

For the purposes of this study, we select WEKA (Waikato Environment for Knowledge Analysis) software that was developed at the University of Waikato in New Zealand**.** WEKA tool supports to a wider range of algorithms & very large data sets. The Weka (pronounced Waykuh) workbench contains a collection of visualization tools & algorithms. WEKA is open source software issued under the GNU General Public License. It contains tools for data pre-processing, classification, regression, clustering, association rules, and visualization. The original non-java version of Weka was a TCL/TK, but the recent java based version is Weka 3(1997), is now used in many different application areas, in particular for education & research.

Weka's main user interface is Explorer. The Experimenter is also there by which we can compare weka's machine learning algorithms' performance. The Explorer interface has many panels by which we can access to main components of workbench. The Visualization tab allows visualizing a 2-D plot of the current working relation, it is very useful. In this study WEKA toolkit 3.6.9 is used for constructing the decision tree.

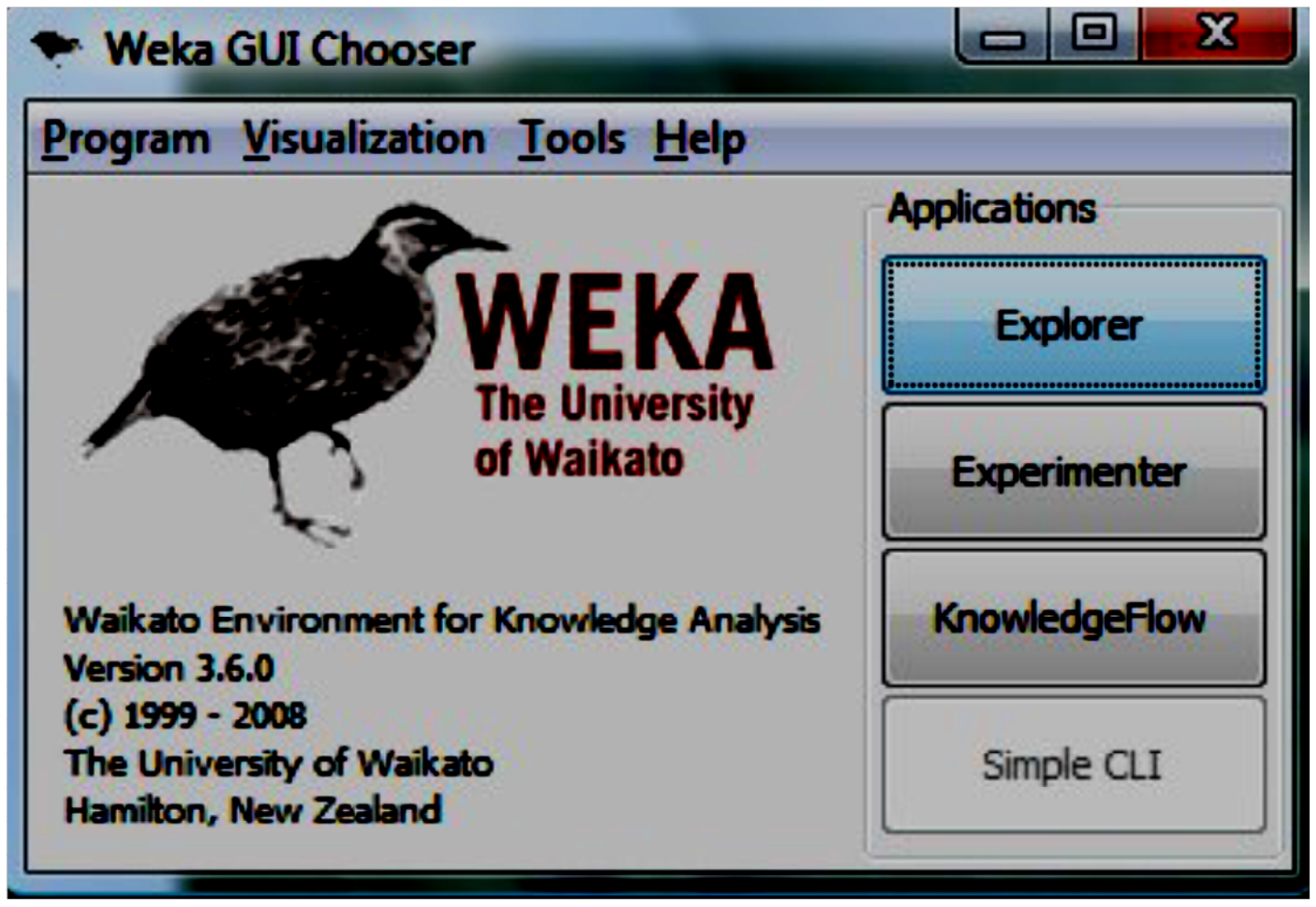

**Figure 3.1: WEKA 3.6.9 Interface**

*SPSS*

SPSS (Statistical Package for the Social Sciences) has now been in development for more than thirty years. Originally developed as a programming language for conducting statistical analysis, it has grown into a complex and powerful application with now uses both a graphical and a syntactical interface and provides dozens of functions for managing, analyzing, and presenting data. SPSS Inc. continues its tradition of regularly enhancing this family of powerful but easy-to-use statistical software products with the release of SPSS 16.0. In this study SPSS 16.0 is used for performing correlation between student's attributes & final grade (PG Grade).

## 3.1.3 Flowchart of process

Step 1: Data Collection
Step 2: Data preprocessing (attribute selection, transformation, normalization)
Step 3: Use the Data mining WEKA tool
Step 4: Apply the ID3 decision tree classification technique on data
Step 5: Evaluate the Results
Step 6: Perform classification on new data using testing
Step 7: Evaluate the accuracy of the model

## 3.1.4 Results

## a) Classification Analysis

After this step, in order to generate the decision tree based on the ID3 algorithm in Weka, we must load our data first, followed by the proper selection of the algorithm from the list provided by this software and then press the Start button.
The output generated has the next sections:

- *Run information*
- *Classifier model*
- *Evaluation on training set*
- *Detailed Accuracy By Class*

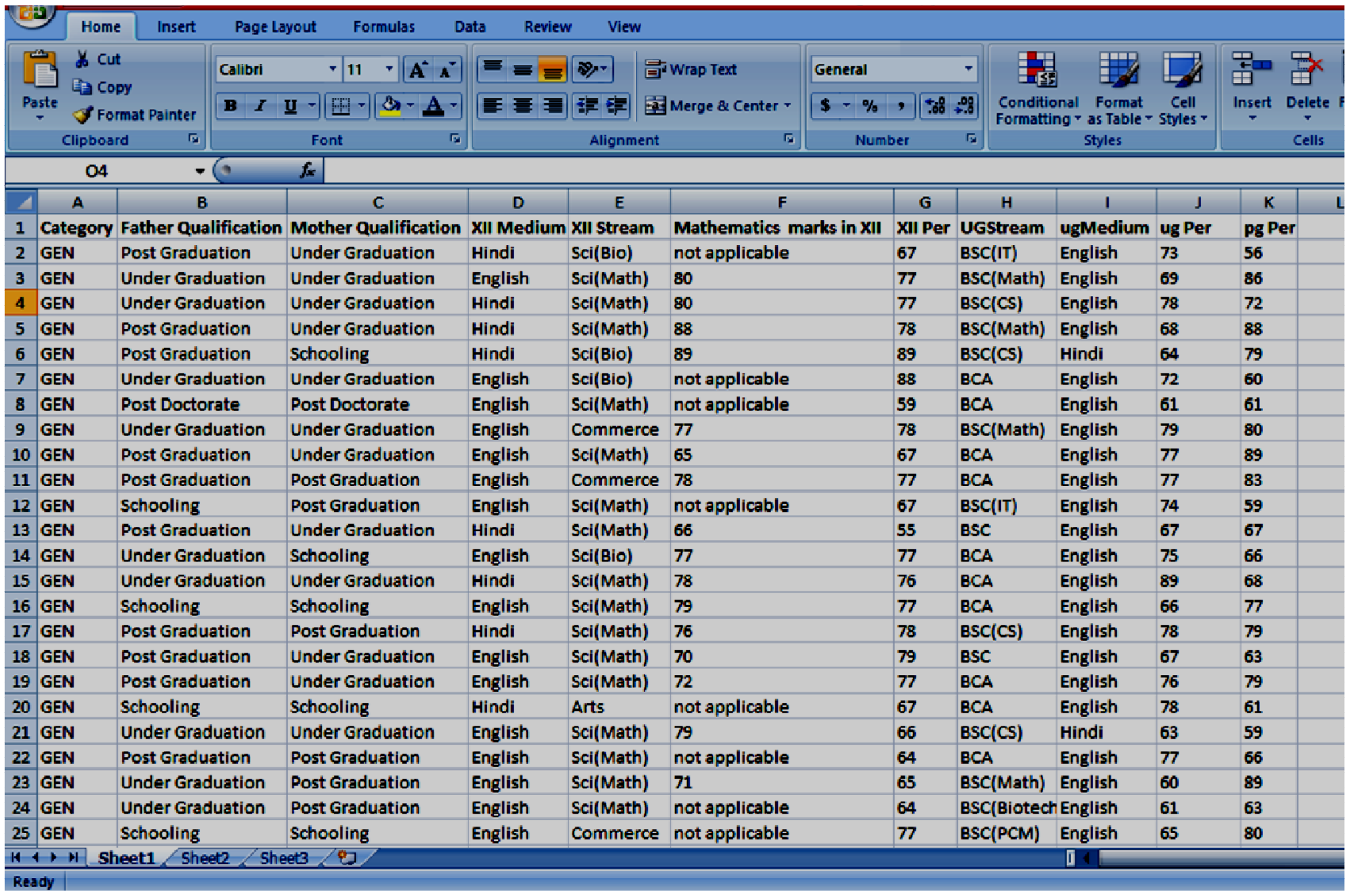

| Category | Father Qualification | Mother Qualification | XII Medium | XII Stream | Mathematics marks in XII | XII Per | UGStream | ugMedium | ug Per | pg Per |
|---|---|---|---|---|---|---|---|---|---|---|
| GEN | Post Graduation | Under Graduation | Hindi | Sci(Bio) | not applicable | 67 | BSC(IT) | English | 73 | 56 |
| GEN | Under Graduation | Under Graduation | English | Sci(Math) | 80 | 77 | BSC(Math) | English | 69 | 86 |
| GEN | Under Graduation | Under Graduation | Hindi | Sci(Math) | 80 | 77 | BSC(CS) | English | 78 | 72 |
| GEN | Post Graduation | Under Graduation | Hindi | Sci(Math) | 88 | 78 | BSC(Math) | English | 68 | 88 |
| GEN | Post Graduation | Schooling | Hindi | Sci(Bio) | 89 | 89 | BSC(CS) | Hindi | 64 | 79 |
| GEN | Under Graduation | Under Graduation | English | Sci(Bio) | not applicable | 88 | BCA | English | 72 | 60 |
| GEN | Post Doctorate | Post Doctorate | English | Sci(Math) | not applicable | 59 | BCA | English | 61 | 61 |
| GEN | Under Graduation | Under Graduation | English | Commerce | 77 | 78 | BSC(Math) | English | 79 | 80 |
| GEN | Post Graduation | Under Graduation | English | Sci(Math) | 65 | 67 | BCA | English | 77 | 89 |
| GEN | Post Graduation | Post Graduation | English | Commerce | 78 | 77 | BCA | English | 77 | 83 |
| GEN | Schooling | Post Graduation | English | Sci(Math) | not applicable | 67 | BSC(IT) | English | 74 | 59 |
| GEN | Post Graduation | Under Graduation | Hindi | Sci(Math) | 66 | 55 | BSC | English | 67 | 67 |
| GEN | Under Graduation | Schooling | English | Sci(Bio) | 77 | 77 | BCA | English | 75 | 66 |
| GEN | Under Graduation | Under Graduation | Hindi | Sci(Math) | 78 | 76 | BCA | English | 89 | 68 |
| GEN | Schooling | Schooling | English | Sci(Math) | 79 | 77 | BCA | English | 66 | 77 |
| GEN | Post Graduation | Post Graduation | Hindi | Sci(Math) | 76 | 78 | BSC(CS) | English | 78 | 79 |
| GEN | Post Graduation | Under Graduation | English | Sci(Math) | 70 | 79 | BSC | English | 67 | 63 |
| GEN | Post Graduation | Under Graduation | English | Sci(Math) | 72 | 77 | BCA | English | 76 | 79 |
| GEN | Schooling | Schooling | Hindi | Arts | not applicable | 67 | BCA | English | 78 | 61 |
| GEN | Under Graduation | Under Graduation | English | Sci(Math) | 79 | 66 | BSC(CS) | Hindi | 63 | 59 |
| GEN | Post Graduation | Post Graduation | English | Sci(Math) | not applicable | 64 | BCA | English | 77 | 66 |
| GEN | Under Graduation | Post Graduation | English | Sci(Math) | 71 | 65 | BSC(Math) | English | 60 | 89 |
| GEN | Under Graduation | Post Graduation | English | Sci(Math) | not applicable | 64 | BSC(Biotech | English | 61 | 63 |
| GEN | Schooling | Schooling | English | Commerce | not applicable | 77 | BSC(PCM) | English | 65 | 80 |

**Figure 3.2: The Weka .csv file for MCA data**

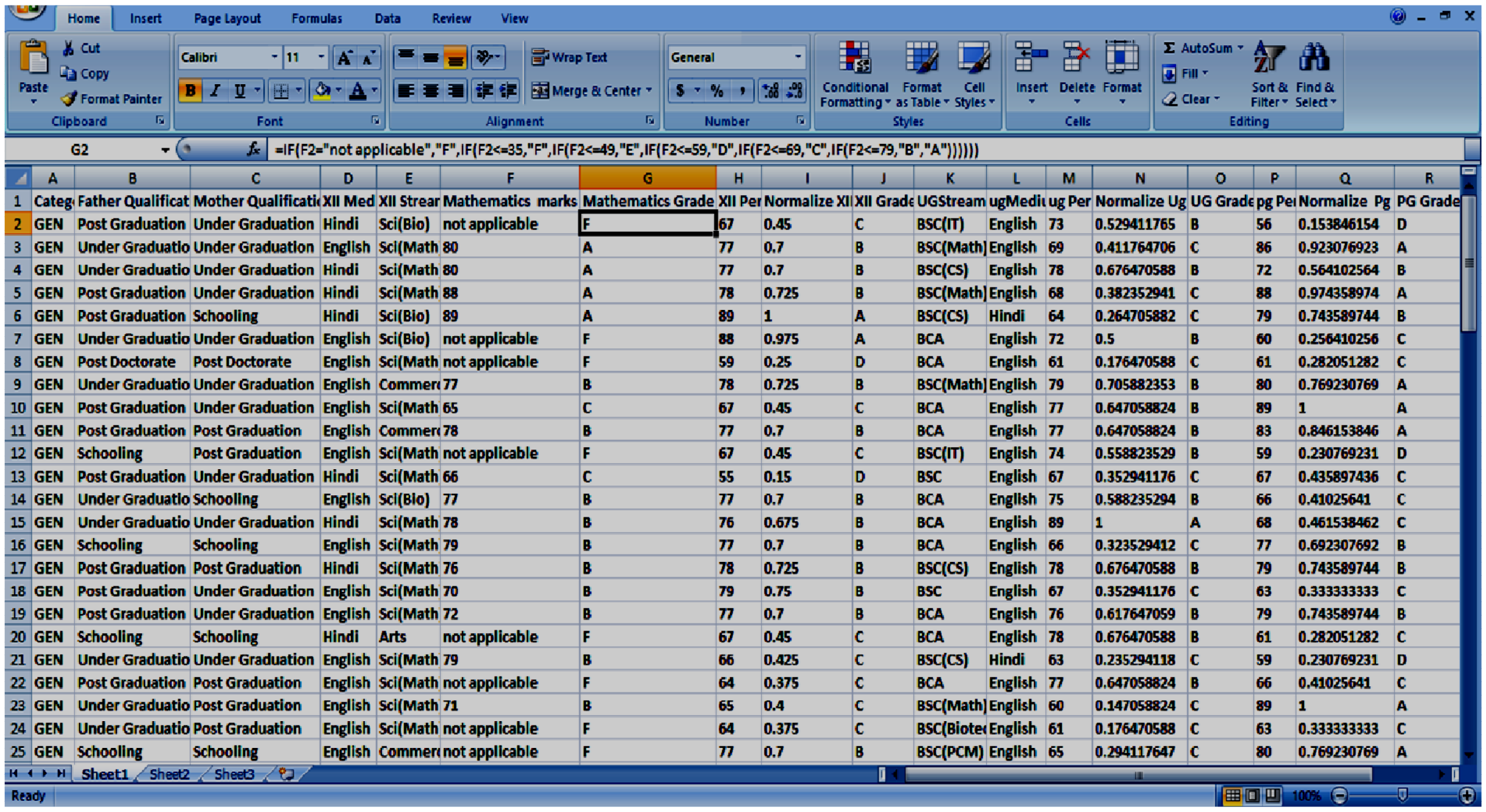

Figure 3.3: The Weka .CSV file for MCA data after the transformation

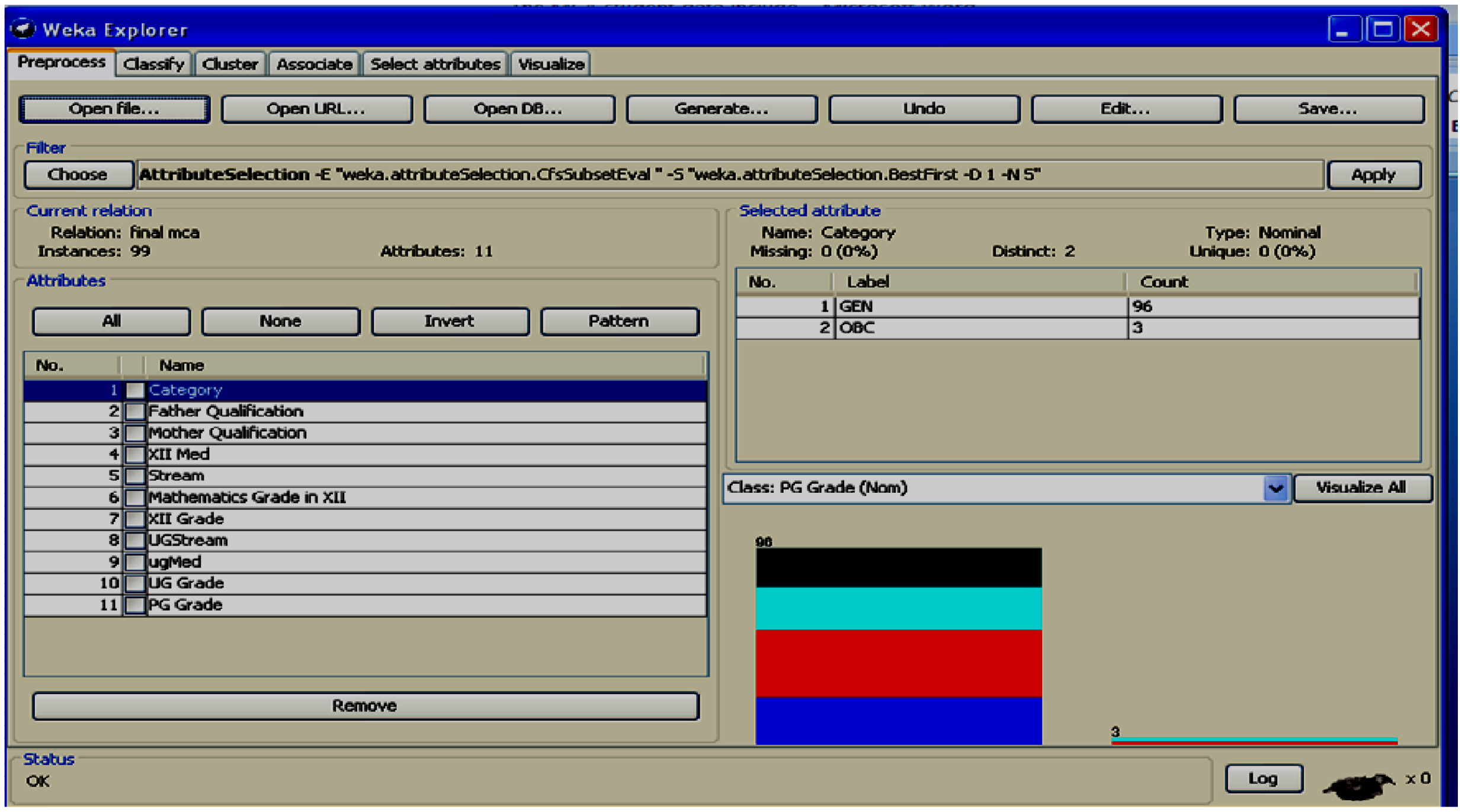

Figure 3.4: Load the final mca.csv file in WEKA

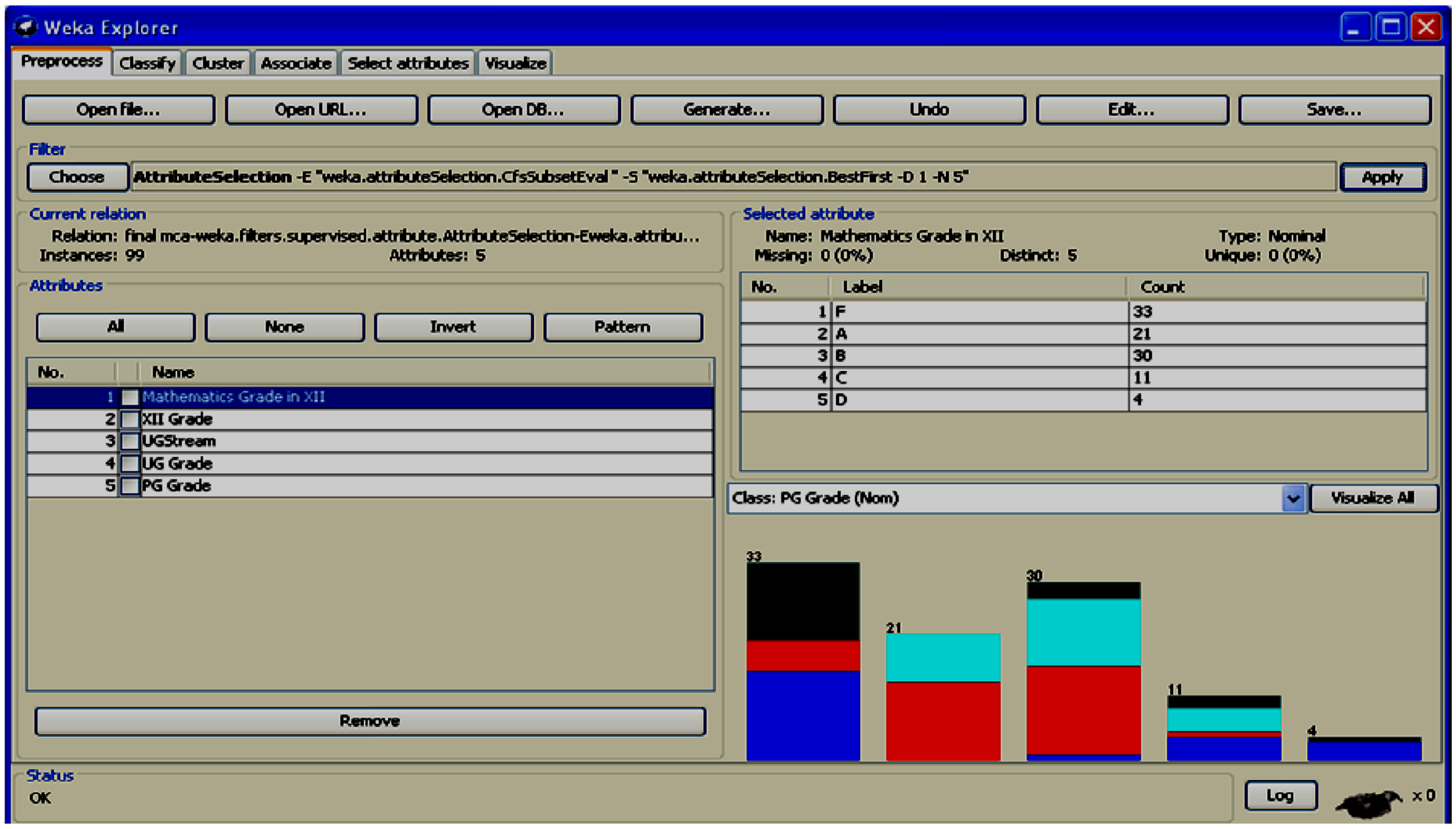

**Figure 3.5: Apply Attribute Selection in the final mca.csv file**

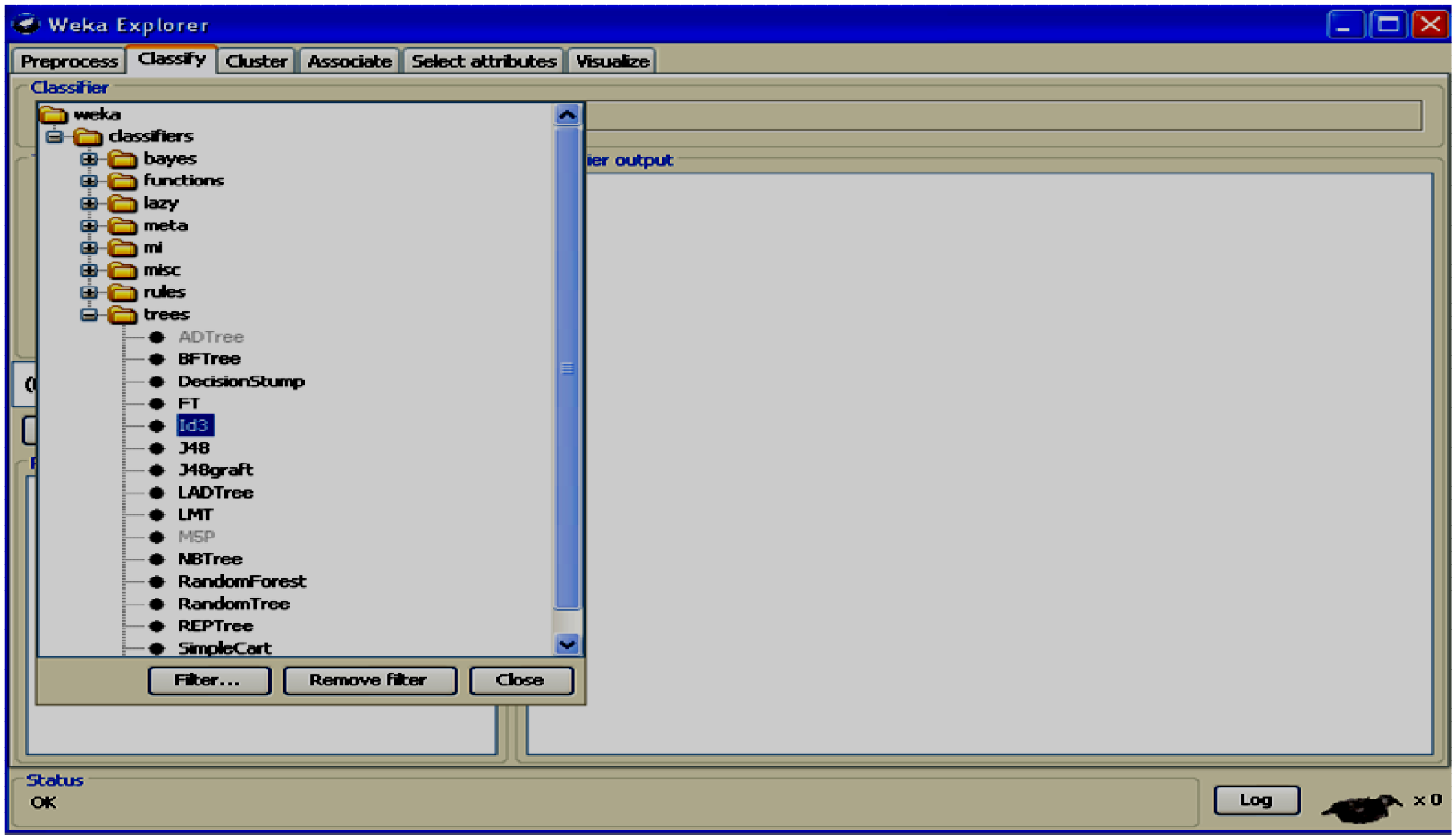

**Figure 3.6: Choose ID3 Decision Tree Classification Algorithm from Classify Tab**

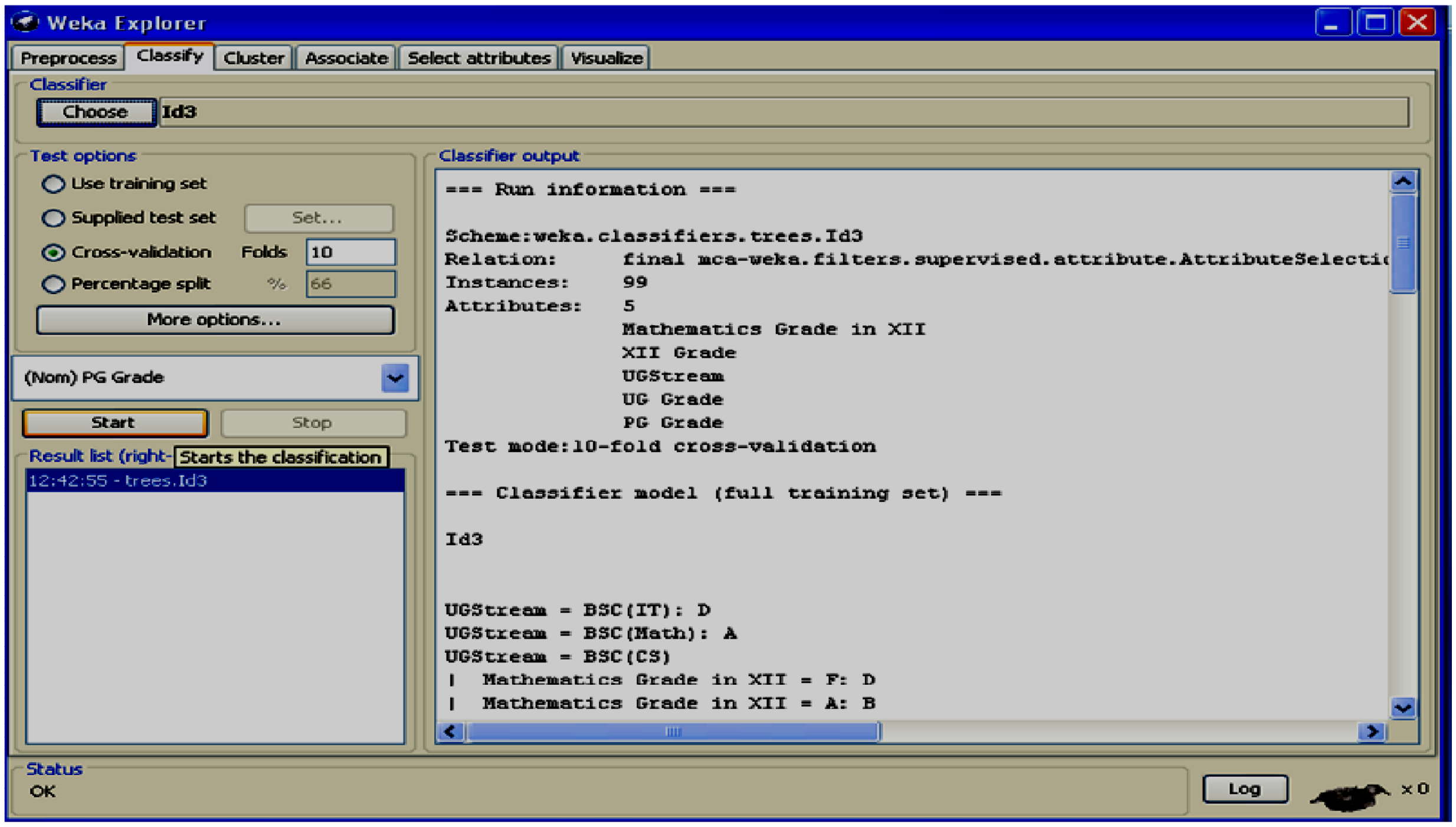

**Figure 3.7: Start the Classification by click on Start Tab**

The *run information* part contains general information about the scheme used, the number of instances (99) and attributes (7) as well as the attributes names as presented in Figure 3.7.

**=== Run information ===**
**Scheme:weka.classifiers.trees.Id3**
**Relation: final mca-weka.filters.supervised.attribute.AttributeSelection-Eweka.attributeSelection.CfsSubsetEval-Sweka.attributeSelection.BestFirst -D 1 -N 5**
**Instances: 99**
**Attributes: 5**
**Mathematics Grade in XII**
**XII Grade**
**UGStream**
**UG Grade**
**PG Grade**
**Test mode:10-fold cross-validation**

**Figure 3.8: The Run Information output**

The second part of the output is represented by the ID3 decision tree (Figure 3.8)
=== Classifier model (full training set) ===

Id3

```
UGStream = BSC(IT): D
UGStream = BSC(Math): A
UGStream = BSC(CS)
|  Mathematics Grade in XII = F: D
|  Mathematics Grade in XII = A: B
|  Mathematics Grade in XII = B
|  |  XII Grade = C: D
|  |  XII Grade = B: B
|  |  XII Grade = A: B
|  |  XII Grade = D: null
|  |  XII Grade = E: null
|  Mathematics Grade in XII = C: B
|  Mathematics Grade in XII = D: null
UGStream = BCA
|  Mathematics Grade in XII = F
|  |  XII Grade = C: C
|  |  XII Grade = B: null
|  |  XII Grade = A
|  |  |  UG Grade = B: C
|  |  |  UG Grade = C: D
|  |  |  UG Grade = A: null
|  |  |  UG Grade = D: null
|  |  XII Grade = D: C
|  |  XII Grade = E: C
|  Mathematics Grade in XII = A: B
|  Mathematics Grade in XII = B
|  |  UG Grade = B: B
|  |  UG Grade = C: B
|  |  UG Grade = A: C
|  |  UG Grade = D: null
|  Mathematics Grade in XII = C
|  |  UG Grade = B: B
|  |  UG Grade = C
|  |  |  XII Grade = C: C
|  |  |  XII Grade = B: null
|  |  |  XII Grade = A: null
|  |  |  XII Grade = D: D
|  |  |  XII Grade = E: null
|  |  UG Grade = A: null
|  |  UG Grade = D: null
|  Mathematics Grade in XII = D
|  |  XII Grade = C: null
|  |  XII Grade = B: C
|  |  XII Grade = A: null
|  |  XII Grade = D: D
```

| | XII Grade = E: null
UGStream = BSC
| XII Grade = C
| | Mathematics Grade in XII = F: C
| | Mathematics Grade in XII = A: null
| | Mathematics Grade in XII = B: null
| | Mathematics Grade in XII = C: D
| | Mathematics Grade in XII = D: null
| XII Grade = B: C
| XII Grade = A: C
| XII Grade = D
| | Mathematics Grade in XII = F: D
| | Mathematics Grade in XII = A: null
| | Mathematics Grade in XII = B: null
| | Mathematics Grade in XII = C: C
| | Mathematics Grade in XII = D: null
| XII Grade = E: null
UGStream = BSC(Biotech): C
UGStream = BSC(PCM): A

Time taken to build model: 0 seconds

**Figure 3.9: The Classifier Model**

Beside this, Weka provides some complementary information about the percent of correctly as well as incorrectly classified instances.
In this example, out of a total of 99 instances, only 67 have been correctly classified meaning 67.6768 %.
This summary is presented in figure 3.9, along with some important statistical parameters.
**Kappa statistic:-**a measure of agreement between two individuals, with a 0.661 value.

**Mean absolute error:-**a quantity used to measure how close forecasts or predictions are to the eventual outcomes.

**Root mean squared error:-**a good measure of the model's accuracy.

**Root relative squared error:-**the average of the actual values.

**Relative absolute error:-**similar to the relative squared error.

**=== Stratified cross-validation ===**
**=== Summary ===**

Correctly Classified Instances 67 67.6768 %

| | | |
|---|---|---|
| Incorrectly Classified Instances | 22 | 22.2222 % |
| Kappa statistic | 0.661 | |
| Mean absolute error | 0.1171 | |
| Root mean squared error | 0.2935 | |
| Relative absolute error | 35.5222 % | |
| Root relative squared error | 72.5685 % | |
| UnClassified Instances | 10 | 10.101 % |
| Total Number of Instances | 99 | |

**Figure 3.10: Evaluation on training set**

The forth part of the output, presented in Figure 3.10, contains information regarding the detailed accuracy by class.

**=== Detailed Accuracy By Class ===**

| | **TP Rate** | **FP Rate** | **Precision** | **Recall** | **F-Measure** | **ROC Area** | **Class** |
|---|---|---|---|---|---|---|---|
| | 0.783 | 0.091 | 0.75 | 0.783 | 0.766 | 0.867 | D |
| | 0.794 | 0.036 | 0.931 | 0.794 | 0.857 | 0.96 | A |
| | 0.889 | 0.141 | 0.615 | 0.889 | 0.727 | 0.835 | B |
| | 0.429 | 0.053 | 0.6 | 0.429 | 0.5 | 0.665 | C |
| **Weighted Average** | 0.753 | 0.074 | 0.768 | 0.753 | 0.751 | 0.864 | |

**=== Confusion Matrix ===**

| **a** | **b** | **c** | **d** | **<-- classified as** |
|---|---|---|---|---|
| 18 | 0 | 1 | 4 | **a = D** |
| 0 | 27 | 7 | 0 | **b = A** |
| 0 | 2 | 16 | 0 | **c = B** |
| 6 | 0 | 2 | 6 | **d = C** |

**Figure 3.11: Detailed Accuracy By Class**

**Accuracy=** (Total number of corrected prediction)/Total number of instances=67/99=**67.67%**

**TP Rate (True positive rate):-**is the proportion of positive cases that were correctly identified, as calculated using the equation:-

**TP=Correctly classified instance/Total instance**

For Class D:- 18/(18+0+1+4)=0.78
For Class A:- 27/(27+7)=0.794
For Class B:- 16/(2+16)=0.888
For Class C:- 6/(6+2+6)=0.428

**FP Rate (False positive rate):-** is the proportion of negatives cases that were incorrectly classified as positive

**Precision(P) :-** is the proportion of the predicted positive cases that were correct, as calculated using the equation:-

**P= Correctly classified instance/Total predicted instance**

For Class D:- 18/(18+6)=0.75
For Class A:- 27/(27+2)=0.93
For Class B:- 16/(1+7+16+2)=0.615
For Class C:- 6/(10+6)=0.6

**Recall:-** TP Rate and Recall are same

**F-measure**:- F-Measure is a measure of a test's accuracy and is determined using the formula:

**(2 * TP Rate * Precision) / (TP Rate + Precision)**

For Class D:-(2*0.783*.75)/(.783+.75)=0.766
For Class A:-(2*0.794*.931)/(.794+.931)=0.85706
For Class B:-(2*0.889*.615)/(.889+.615)=0.727
For Class C:-(2*0.429*.6)/(.429+.6)=0.5

**ROC (Receiver Operating Characteristic Area) –** The ROC curve is given by the TP Rate and FP Rate. The area under the ROC Curve (AUC) is a method of measuring the performance of the ROC curve. If AUC is 1 then the prediction is perfect; if it is 0.5 then the prediction is random. Analyzing our output we conclude that even if the prediction in this case is not perfect, it is not random as well. The best prediction is **for class A – 0.96** and the "weak" prediction is **0.665 for class C**. Between this extremes are the values 0.867 for class D and 0.835 for class B.

**Weighted Average:-**
An average in which each quantity to be averaged is assigned a weight. These weightings determine the relative importance of each quantity on the average.

**For TP Rate:-**
**Sum((TP Rate of Class* number of instance with that class))/Sum(number of instances with that class)**

=((0.783*23)+(.794*34)+(.889*18)+(.429*14))/89=0.7529

**For FP Rate:-**
**Sum((FP Rate of Class* number of instance with that class))/Sum(number of instances with that class)**

=((0.091*23)+(.036*34)+(.141*18)+(.053*14))/89=0.0741

**For Precision:-**
**Sum((Precision of Class* number of instance with that class))/Sum(number of instances with that class)**

=((0.75*23)+(.931*34)+(.615*18)+(.6*14))/89=0.768

**For Recall:-**
**Sum((Recall of Class* number of instance with that class))/Sum(number of instances with that class)**

=((0.783*23)+(.794*34)+(.889*18)+(.429*14))/89=0.7529

**For F-Measure:-**
**Sum((F-Measure of Class* number of instance with that class))/Sum(number of instances with that class)**

=((0.766*23)+(.857*34)+(.727*18)+(.5*14))/89=0.751

**For ROC Area:-**
**Sum((ROC Area of Class* number of instance with that class))/Sum(number of instances with that class)**=((0.867*23)+(.96*34)+(.835*18)+(.665*14))/89=0.86428

# b) Correlation Analysis

**Table 3.1: Correlation between Mathematics Grade in XII & PG Grade**

**Correlations**

| | | Mathematics Grade in XII | PG Grade |
|---|---|---|---|
| Mathematics Grade in XII | Pearson Correlation | 1 | .629** |
| | Sig. (2-tailed) | | .000 |
| | N | 99 | 99 |
| PG Grade | Pearson Correlation | .629** | 1 |
| | Sig. (2-tailed) | .000 | |
| | N | 99 | 99 |

**. Correlation is significant at the 0.01 level (2-tailed).

| | |
|---|---|
| Mathematics Grade in XII with Mathematics Grade in XII | Mathematics Grade in XII with PG Grade |
| PG Grade with Mathematics Grade in XII | PG Grade with PG Grade |

In the matrix, there are four correlations
"Mathematics Grade in XII with Mathematics Grade in XII" and "PG Grade with PG Grade" will be of course 1 (perfect positive correlation).

**Mathematics Grade in XII with PG Grade and PG Grade with Mathematics Grade in XII correlations tell us three things:**

- The Pearson correlation is .629**, as there is no minus sign preceding the correlation coefficient means that the relationship between XII Grade and PG Grade is positive. There are two asterisks after the correlation coefficient means correlation is significant at the 0.01 level(2-tailed).

- The significance level or p-value is .000, low value indicate a low probability of finding a relationship between these variables.
- N is the number of cases that contains data and this is 99.

**Table 3.2: Correlation between UG Grade & PG Grade**

**Correlations**

| | | UG Grade | PG Grade |
|---|---|---|---|
| UG Grade | Pearson Correlation | 1 | .498** |
| | Sig. (2-tailed) | | .000 |
| | N | 99 | 99 |
| PG Grade | Pearson Correlation | .498** | 1 |
| | Sig. (2-tailed) | .000 | |
| | N | 99 | 99 |

**. Correlation is significant at the 0.01 level (2-tailed).

**Table 3.3: Correlation between UG Stream & PG Grade**

**Correlations**

| | | UGStream | PG Grade |
|---|---|---|---|
| UG Stream | Pearson Correlation | 1 | .443** |
| | Sig. (2-tailed) | | .000 |
| | N | 99 | 99 |
| PG Grade | Pearson Correlation | .443** | 1 |
| | Sig. (2-tailed) | .000 | |
| | N | 99 | 99 |

**. Correlation is significant at the 0.01 level (2-tailed).

**Table 3.4: Correlation Between X Per & PG Grade**

**Correlations**

| | | X Per | PG Grade |
|---|---|---|---|
| X Per | Pearson Correlation | 1 | .290** |
| | Sig. (2-tailed) | | .004 |
| | N | 99 | 99 |
| PG Grade | Pearson Correlation | .290** | 1 |
| | Sig. (2-tailed) | .004 | |
| | N | 99 | 99 |

**. Correlation is significant at the 0.01 level (2-tailed).

**Table 3.5: Correlation Between XII Grade & PG Grade**

**Correlations**

| | | XII Grade | PG Grade |
|---|---|---|---|
| XII Grade | Pearson Correlation | 1 | .124 |
| | Sig. (2-tailed) | | .222 |
| | N | 99 | 99 |
| PG Grade | Pearson Correlation | .124 | 1 |
| | Sig. (2-tailed) | .222 | |
| | N | 99 | 99 |

**XII Grade with PG Grade and PG Grade with XII Grade correlations tell us three things:**

- The Pearson correlation is .124, as there is no minus sign preceding the correlation coefficient means that the relationship between XII Grade and PG Grade is positive. In other words, students who score high in XII Grade will tend to score high in PG Grade.
- The significance level or p-value is .222.
- N is 99 (the number of cases that contains data).

**Table 3.6: Correlation Between UG Med & PG Grade**

**Correlations**

| | | UG Med | PG Grade |
|---|---|---|---|
| UG Med | Pearson Correlation | 1 | .118 |
| | Sig. (2-tailed) | | .246 |
| | N | 99 | 99 |
| PG Grade | Pearson Correlation | .118 | 1 |
| | Sig. (2-tailed) | .246 | |
| | N | 99 | 99 |

# c) Cross Tabulation Analysis

| PG Grade | UG Medium(English) | UG Medium(Hindi) |
|---|---|---|
| A | 27 | 7 |
| B | 18 | 5 |
| C | 18 | 1 |
| D | 20 | 3 |

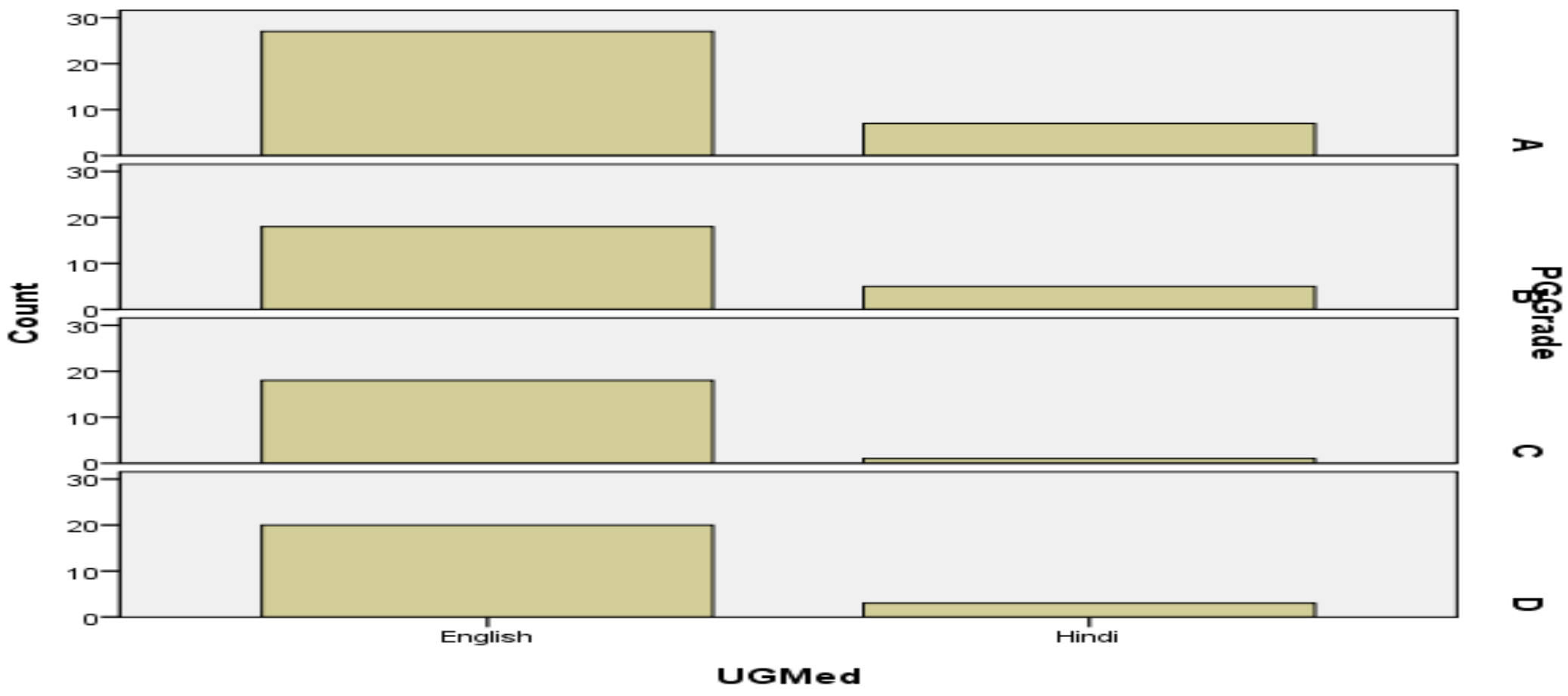

**Figure 3.12: Cross Tabulation between PG Grade & UG Medium**

**Table 3.7: UG Med * PG Grade Cross tabulation**

| Count | | | | | | |
|---|---|---|---|---|---|---|
| | | PG Grade | | | | Total |
| | | A | B | C | D | |
| UG Med | English | 27 | 18 | 18 | 20 | 83 |
| | Hindi | 7 | 5 | 1 | 3 | 16 |
| Total | | 34 | 23 | 19 | 23 | 99 |

**MCA PG Grade with University Standard**

| PG Grade | University Std (Good) | University Std (Very Good) | University Std (Excellent) |
|---|---|---|---|
| A | 10 | 16 | 8 |
| B | 8 | 10 | 5 |
| C | 7 | 8 | 4 |
| D | 13 | 5 | 5 |

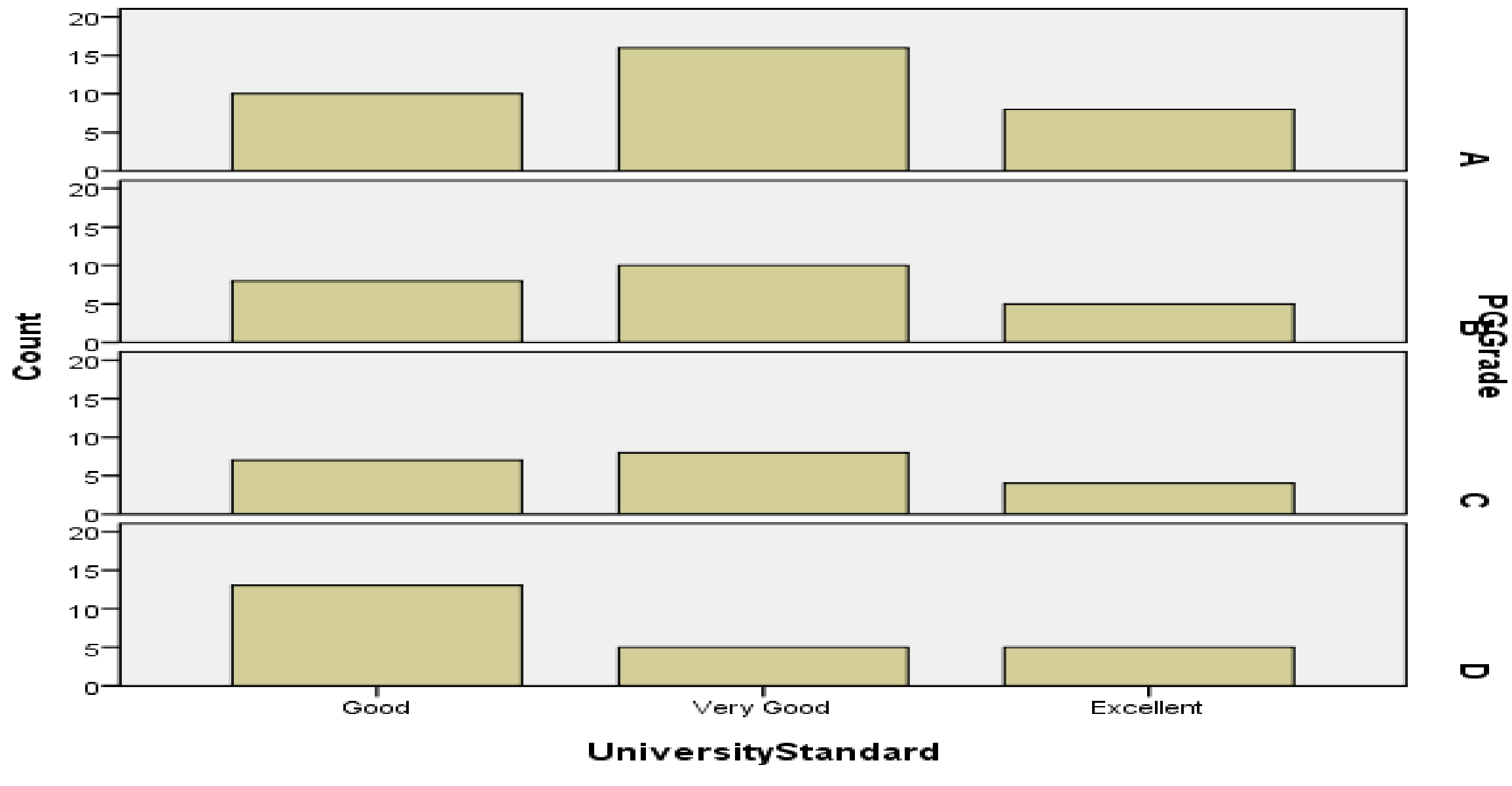

**Figure 3.13: Cross Tabulation between PG Grade & University Standard**

**Table 3.8: University Standard * PG Grade Cross Tabulation**

Count

| | | PG Grade | | | | |
|---|---|---|---|---|---|---|
| | | A | B | C | D | Total |
| University Standard | Good | 10 | 8 | 7 | 13 | 38 |
| | Very Good | 16 | 10 | 8 | 5 | 39 |
| | Excellent | 8 | 5 | 4 | 5 | 22 |
| Total | | 34 | 23 | 19 | 23 | 99 |

**MCA PG Grade with Under Graduation Stream**

| PG Grade | University Stream BSC(Math) | University Stream BCA | University Stream BSC |
|---|---|---|---|
| A | 27 | 7 | 0 |
| B | 10 | 13 | 0 |
| C | 0 | 13 | 6 |
| D | 11 | 5 | 7 |

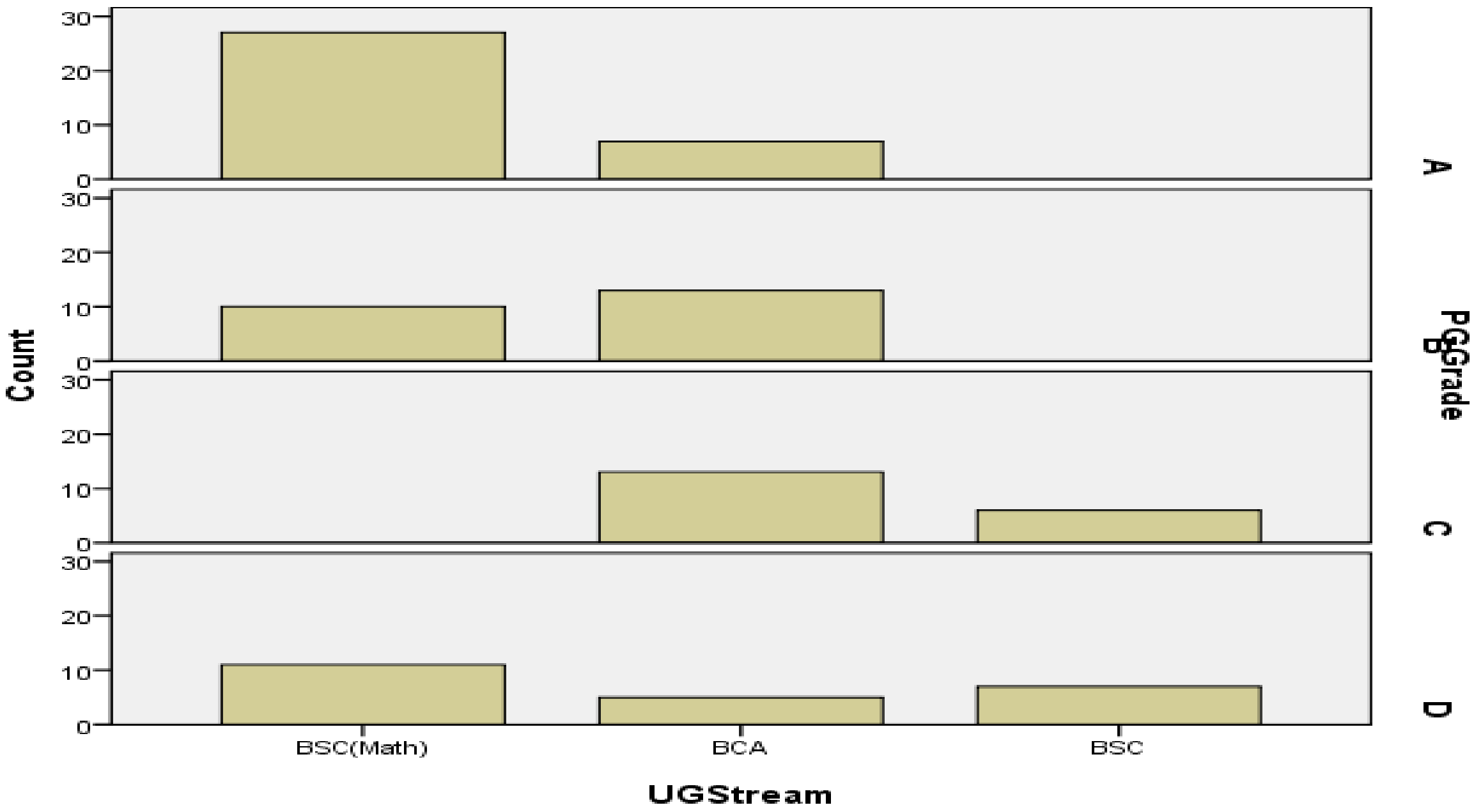

**Figure 3.14: Cross Tabulation between PG Grade & UG Stream**

## MCA PG Grade with XII Medium

| PG Grade | XII Medium-English | XII Medium-Hindi |
|---|---|---|
| A | 16 | 18 |
| B | 16 | 7 |
| C | 8 | 11 |
| D | 14 | 9 |

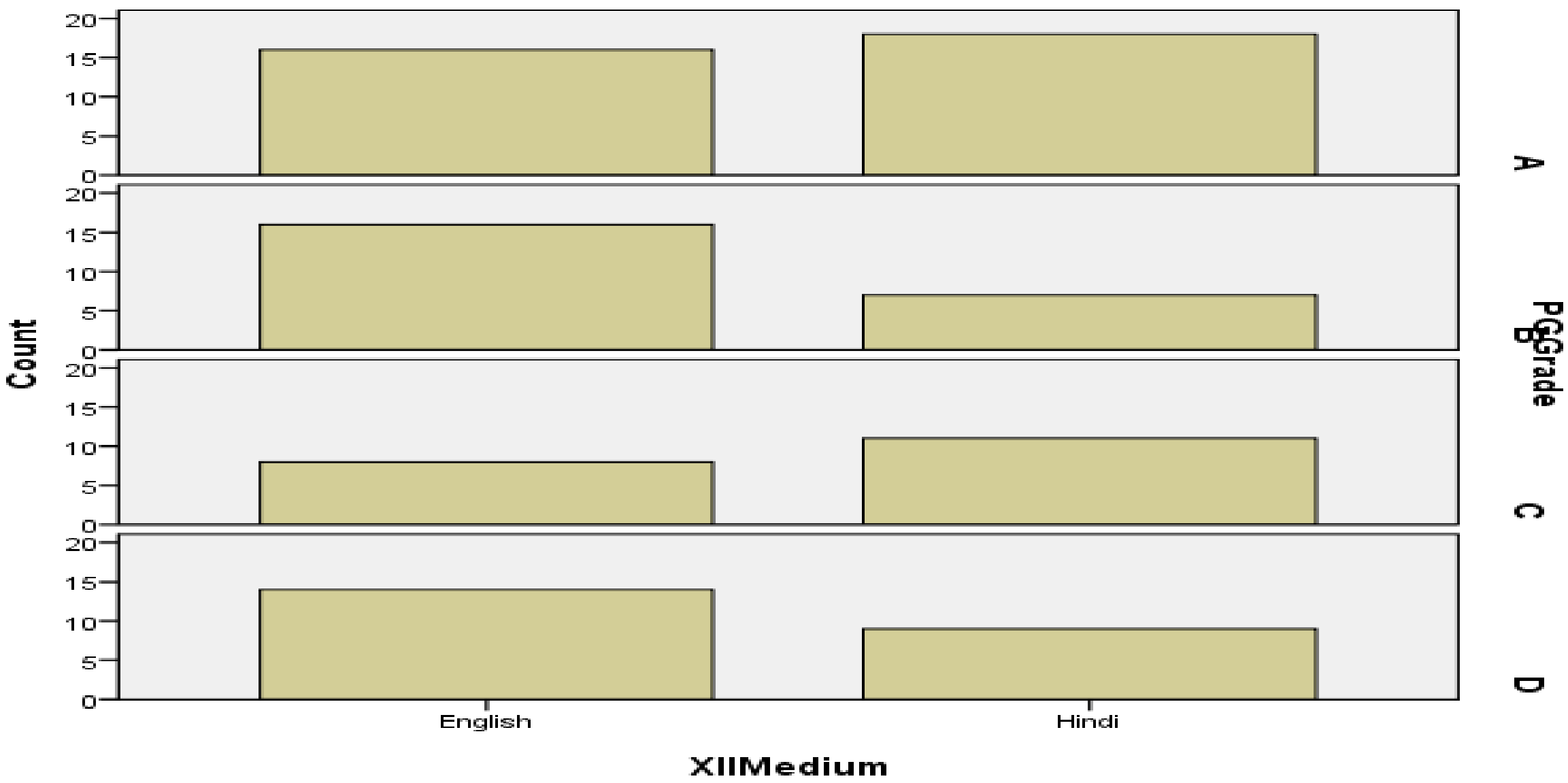

**Figure 3.15: Cross Tabulation between PG Grade & XII Medium**

**Table 3.9: XII Medium * PG Grade Cross Tabulation**

Count

| | | PG Grade | | | | Total |
|---|---|---|---|---|---|---|
| | | A | B | C | D | |
| XII Medium | English | 16 | 16 | 8 | 14 | 54 |
| | Hindi | 18 | 7 | 11 | 9 | 45 |
| Total | | 34 | 23 | 19 | 23 | 99 |

## d) ID3 Decision Tree Creation

*Steps of ID3 decision tree creation for MCA student performance*

*Step 1:* **the dataset S of 99 instances with grades 34 "A", 23 "B", 19 "C" and 23 "D".**

**Entropy (S)** = -(34/99) Log2 (34/99) – (23/99) Log2 (23/99) -(19/99) Log2 (19/99) – (23/99) Log2 (23/99) = 1.965031

*STEP 2:* **Attribute XII Grade**

XII Grade value can be A,B,C,D and E.

XII Grade=A is of occurrence 17 (3 of the examples are "B", 3 of the examples are "C",11 of the examples are "D"

XII Grade=B is of occurrence 48 ( 28-A, 16-B, 4-C)

XII Grade=C is of occurrence 23 (6-A, 4-B, 6-C, 7-D)

XII Grade=D is of occurrence 10 (5-C,5-D)

XII Grade=E is of occurrence 1 (1-C)

**Entropy(SXA)=** -(3/17) x log2(3/17) – (3/17) x log2(3/17) –(11/17) x log2(11/17) =1.289609

**Entropy(SXB)=** -(28/48) x log2(28/48) – (16/48) x log2(16/48) –(4/48) x log2(4/48) =1.280672

**Entropy(SXC)=** -(6/23) x log2(6/23) –(4/23) x log2(4/23)–(6/23) x log2(6/23) –(7/23) x log2(7/23)= 1.972647

**Entropy (SXD) =** - (5/10) x log2 (5/10) – (5/10) x log2 (5/10) =1

**Entropy (SXE) =** - (1/1) x log2 (1/1) = 0

**Gain (S, XII Grade)** = Entropy (S) – (17/99) x Entropy (SXA) - (48/99) x Entropy (SXB) - (23/99) x Entropy (SXC)-(10/99) x Entropy (SXD) - (1/99) x Entropy (SXE)

**Gain(S,XII Grade)=** 1.965031-(17/99) x 1.289609- (48/99) x 1.280672- (23/99) x 1.972647 - (10/99) x 1- (1/99) x 0=0.563349283

*STEP 3:* **Attribute Mathematic Grade in XII**

**Its value can be A,B,C,D,E and F**

Mathematics Grade in XII =A is of occurrence 21 (13-A,8-B)

Mathematics Grade in XII =B is of occurrence 29 (14-A, 11-B, 3-C, 1-D)

Mathematics Grade in XII =C is of occurrence 12 (2-A, 4-B, 2-C, 4-D)

Mathematics Grade in XII =D is of occurrence 4 (1-C, 3-D )

Mathematics Grade in XII =E is of occurrence 0

Mathematics Grade in XII =F is of occurrence 33 (5-A, 13-C, 15-D)

**Entropy(SMA)=** -(13/21) x log2(13/21) – (8/21) x log2(8/21)= 0.958711

**Entropy(SMB)=** -(14/29) x log2(14/29) – (11/29) x log2(11/29) -(3/29) x log2(3/29) – (1/29) x log2(1/29) =1.543787

**Entropy(SMC)=** -(2/12) x log2(2/12) – (4/12) x log2(4/12) -(2/12) x log2(2/12) – (4/12) x log2(4/12)= 1.918295

**Entropy (SMD)=** -(1/4) x log2(1/4) – (3/4) x log2(3/4) =0.811278

**Entropy(SMF)=** -(5/33) x log2(5/33) – (13/33) x log2(13/33)- (15/33) x log2(15/33)= 1.458978

**Gain (S, Mathematics Grade in XII)** = Entropy (S) – (21/99) x Entropy (SMA) - (29/99) x Entropy (SMB) - (12/99) x Entropy (SMC)-(4/99) x Entropy (SMD) - (33/99) x Entropy (SMF)

**Gain (S, Mathematics Grade in XII)** = 1.965031 – (21/99) x 0.958711- (29/99) x 1.543787- (12/99) x 1.918295-(4/99) x 0.811278- (33/99) x 1.458978=0.557822111

*STEP 4:* **Attribute UG Stream**

**Its value can be BSC(Math),BSC(PCM),BSC(IT),BSC(CS),BSC(Biotech),BSC,BCA**

UG Stream =BSC(Math) is of occurrence 23 (23-A)

UG Stream =BSC(PCM) is of occurrence 4 (4-A)

UG Stream =BSC(IT) is of occurrence 2 (2-D)

UG Stream =BSC(CS) is of occurrence 21 (10-B, 11-D)

UG Stream =BSC(Biotech) is of occurrence 1 (1-C)

UG Stream =BSC is of occurrence 10 (5-C,5-D)

UG Stream =BCA is of occurrence 38 (7-A, 13-B, 13-C, 5-D)

**Entropy (SBSC (M)) =** -(23/23)xlog2(23/23)=0

**Entropy (SBSC (PCM)) =** -(4/4)xlog2(4/4)=0

**Entropy (SBSC (IT)) =** -(2/2)xlog2(2/2)=0

**Entropy (SBSC (CS)) =** -(10/21)xlog2(10/21) -(11/21)xlog2(11/21)= 0.998364

**Entropy (SBSC(Biotech))=** -(1/1)xlog2(1/1) =0

**Entropy (SBSC) =** -(5/10)xlog2(5/10) -(5/10)xlog2(5/10)= 1

**Entropy (SBCA) =** -(7/38)xlog2(7/38) -(13/38)xlog2(13/38) -(13/38)xlog2(13/38) -(5/38)xlog2(5/38)= 1.893387

**Gain (S,UG Stream)**= Entropy (S)-(23/99) x Entropy (SBSC(M)) - (4/99) x Entropy (SBSC(PCM)) - (2/99) x Entropy (BSC(IT))-(21/99) x Entropy (BSC(CS))-(1/99) x Entropy (SBSC(Biotech))-(10/99) x Entropy (SBSC)- (38/99) x Entropy (SBCA)

**Gain (S,UG Stream)**= 1.965031-(23/99) x 0-(4/99) x 0 - (2/99) x 0 -(21/99) x 0.998364-(1/99) x 0 -(10/99) x1- (38/99) x 1.893387=0.925492111

*STEP 5:* **Attribute UG Grade**

**Its value can be A , B, C, D and E.**

UG Grade =A is of occurrence 2 (1-A,1-C)

UG Grade =B is of occurrence 44 (19-A,19-B,4-C,2-D)

UG Grade =C is of occurrence 43 (14-A,3-B,13-C,13-D)

UG Grade =D is of occurrence 10 (1-B,1-C,8-D)

**Entropy (SUA)=** -(1/2)xlog2(1/2) -(1/2)xlog2(1/2)= 1

**Entropy (SUB)=** -(19/44)xlog2(19/44) -(19/44)xlog2(19/44) -(4/44)xlog2(4/44) -(2/44)xlog2(2/44)= 1.563494

**Entropy (SUC)=** -(14/43)xlog2(14/43) -(3/43)xlog2(3/43) -(13/43)xlog2(13/43) -(13/43)xlog2(13/43)= 1.838607

**Entropy (SUD)=** -(1/10)xlog2(1/10) -(1/10)xlog2(1/10) -(8/10)xlog2(8/10)= 0.921928

**Gain (S,UG Grade)**= Entropy (S)-(2/99) x Entropy (SUA) - (44/99) x Entropy (SUB) - (43/99) x Entropy (SUC)-(10/99) x Entropy (SUD)

**Gain (S,UG Grade)**= 1.965031-(2/99) x1- (44/99) x 1.563494- (43/99) x 1.838607-(10/99) x 0.921928=0.35823184

*STEP 6:* **Find which attribute is the root node and rank the attribute according to information gain shown in Table**

| | |
|---|---|
| **Gain (S,UG Stream)** | **0.925492111** |
| **Gain (S, XII Grade)** | 0.563349283 |
| **Gain(S, Mathematics Grade in XII)** | 0.557822111 |
| **Gain (S,UG Grade)** | 0.35823184 |

**Gain (S, UG Stream)= 0.925492111 is highest**

**Therefore UG Stream is root node**

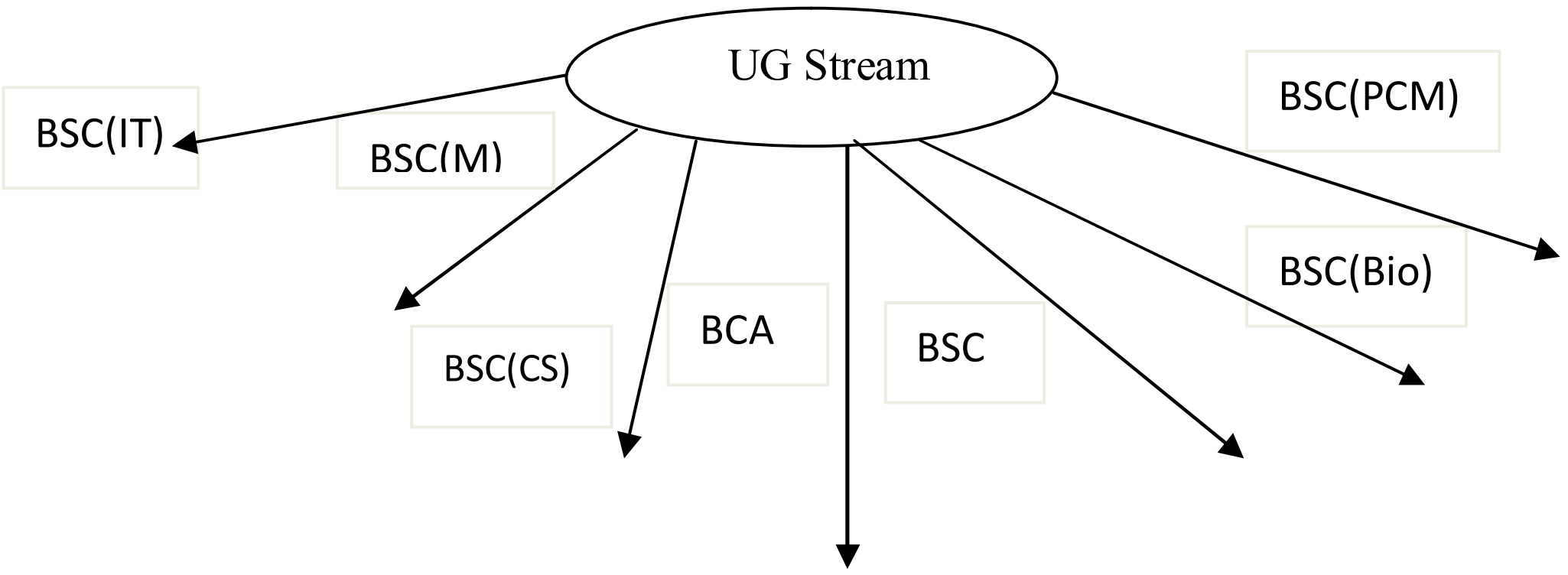

*STEP 7:* **Find which attribute is next decision node.**

Since Entropy (SBSC(M))=0 so no classification

UG Stream->BSC(Math):A (No further classification)

Since Entropy (SBSC(IT))=0 so no classification

UG Stream->BSC (IT):D (No further classification)

Since Entropy (SBSC(Biotech))=0 so no classification

UG Stream->BSC(Biotech):C (No further classification)

Since Entropy (SBSC(PCM))=0 so no classification

UG Stream->BSC(PCM):A (No further classification)

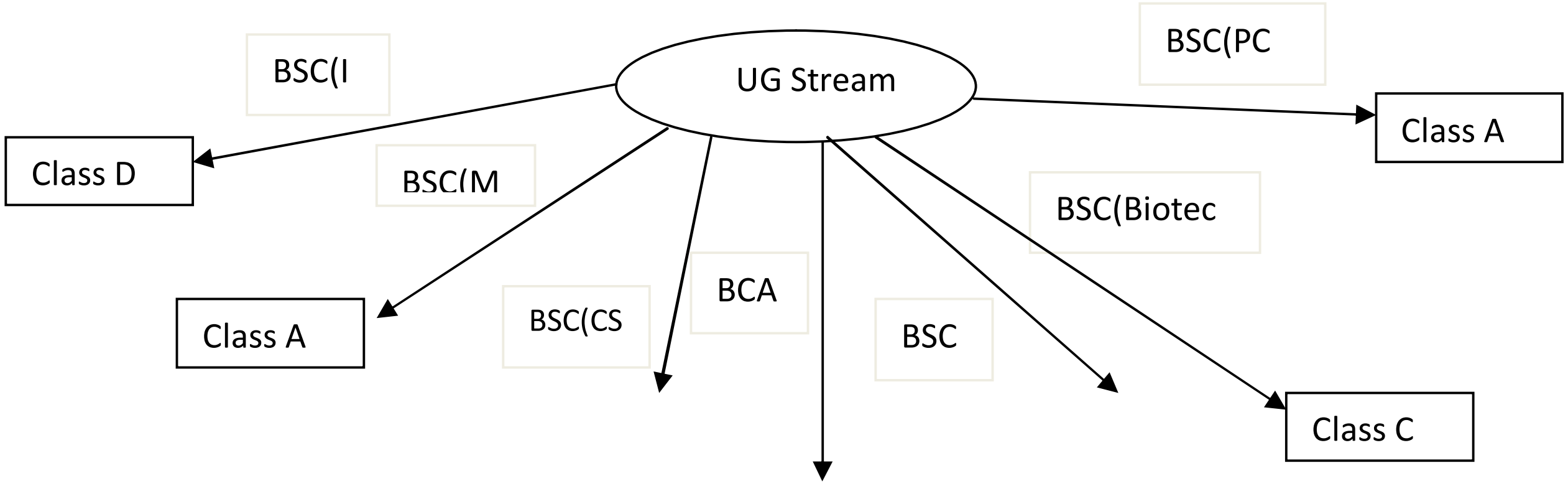

Since

**Entropy (SBSC (CS)) =** 0.998364

**Entropy (SBSC) =** -(5/10)xlog2(5/10) -(5/10)xlog2(5/10)= 1

**Entropy (SBCA)** = -(7/38)xlog2(7/38) -(13/38)xlog2(13/38) -(13/38)xlog2(13/38) -(5/38)xlog2(5/38)= 1.893387

**With UG Stream= BSC (CS)**

1 .For finding the next node in branch of BSC (CS) , calculate the gain value of BSC(CS) with other nodes (XII Grade, Mathematics Grade in XII and UG Grade)

a) Gain(SBSC(CS)),XII Grade) = ?

XII Grade=A is of occurrence 13(0-A,3-B, 0-C, 10-D)

XII Grade=B is of occurrence 6 (0-A, 6-B)

XII Grade=C is of occurrence 2(1-D,1-B)

XII Grade=D is of occurrence 0

**Entropy (SBSC(CS)XA)=** -(3/13)xlog2(3/13) –(10/13)xlog2(10/13)= 0.77934984

**Entropy (SBSC(CS)XB)=** -(6/6)xlog2(6/6)=0

**Entropy (SBSC(CS)XC)=** -(1/2)xlog2(1/2) –(1/2)xlog2(1/2)= 1

Gain(SBSC(CS)),XIIGrade) = Entropy(SBSC(CS))-(13/21)*Entropy(**SBSC(CS)XA)-(6/21)* Entropy (SBSC(CS)XB)-(2/21)* Entropy (SBSC(CS)XC)**

Gain(SBSC(CS)),XIIGrade)= 0.998364 –(13/21)* 0.779349**-(6/21)*0-(2/21)* 1=**0.42067176

b) Gain (SBSC (CS)), Mathematics Grade in XII) = ?

Mathematics Grade in XII =A is of occurrence 7(7-B)

Mathematics Grade in XII =B is of occurrence 3(2-B,1-D)

Mathematics Grade in XII =C is of occurrence 1(1-B)

Mathematics Grade in XII =F is of occurrence 10(10-D)

**Entropy (SBSC(CS)MA)=** -(7/7)xlog2(7/7) =0

**Entropy (SBSC(CS)MB)=** -(2/3)xlog2(2/3)-(1/3)xlog2(1/3) =0.918296

**Entropy (SBSC(CS)MC)=** -(1/1)xlog2(1/1) =0

**Entropy (SBSC(CS)MF)=** -(10/10)xlog2(10/10) =0

**Gain(SBSC(CS)),Mathematics Grade in XII)=**

Entropy(SBSC(CS))-(7/21)*Entropy(SBSC(CS)MA)-(3/21)* Entropy (SBSC(CS)MB)-(1/21)* Entropy (SBSC(CS)MC)-(10/21)*Entropy(SBSC(CS)MF)

Gain(SBSC(CS)),Mathematics Grade in XII) =0.998364-(7/21)*0**-(3/21)*** 0.918296**-(1/21)*0-(10/21)*0=**0.867179

**c) Gain (SBSC (CS), UG Grade) = ?**

UG Grade =A is of occurrence 0

UG Grade =B is of occurrence 7(7-B)

UG Grade =C is of occurrence 7 (2-B,5-D)

UG Grade =D is of occurrence 7(1-B,6-D)

**Entropy (SBSC(CS)UA)=** 0,**Entropy (SBSC(CS)UB)=** 0

**Entropy (SBSC(CS)UC)=** -(2/7)xlog2(2/7)-(5/7)xlog2(5/7)= 0.863120569

**Entropy (SBSC(CS)UD)=** -(1/7)xlog2(1/7)-(6/7)xlog2(6/7)= 0.591672779

Gain(SBSC(CS),UG Grade) =Entropy(SBSC(CS))-(7/21)*Entropy(**SBSC(CS)UB)-(7/21)* Entropy (SBSC(CS)UC)-(7/21)* Entropy (SBSC(CS)UD)**

Gain(SBSC(CS),UG Grade)=

0.998364-(7/21)*0**-(7/21)***0.863120569**-(7/21)*** 0.591672779=0.513432884

**Gain(SBSC(CS),Mathematics Grade in XII) is greater than others attribute, so Mathematics Grade in XII is the next node in BSC(CS) branch.**

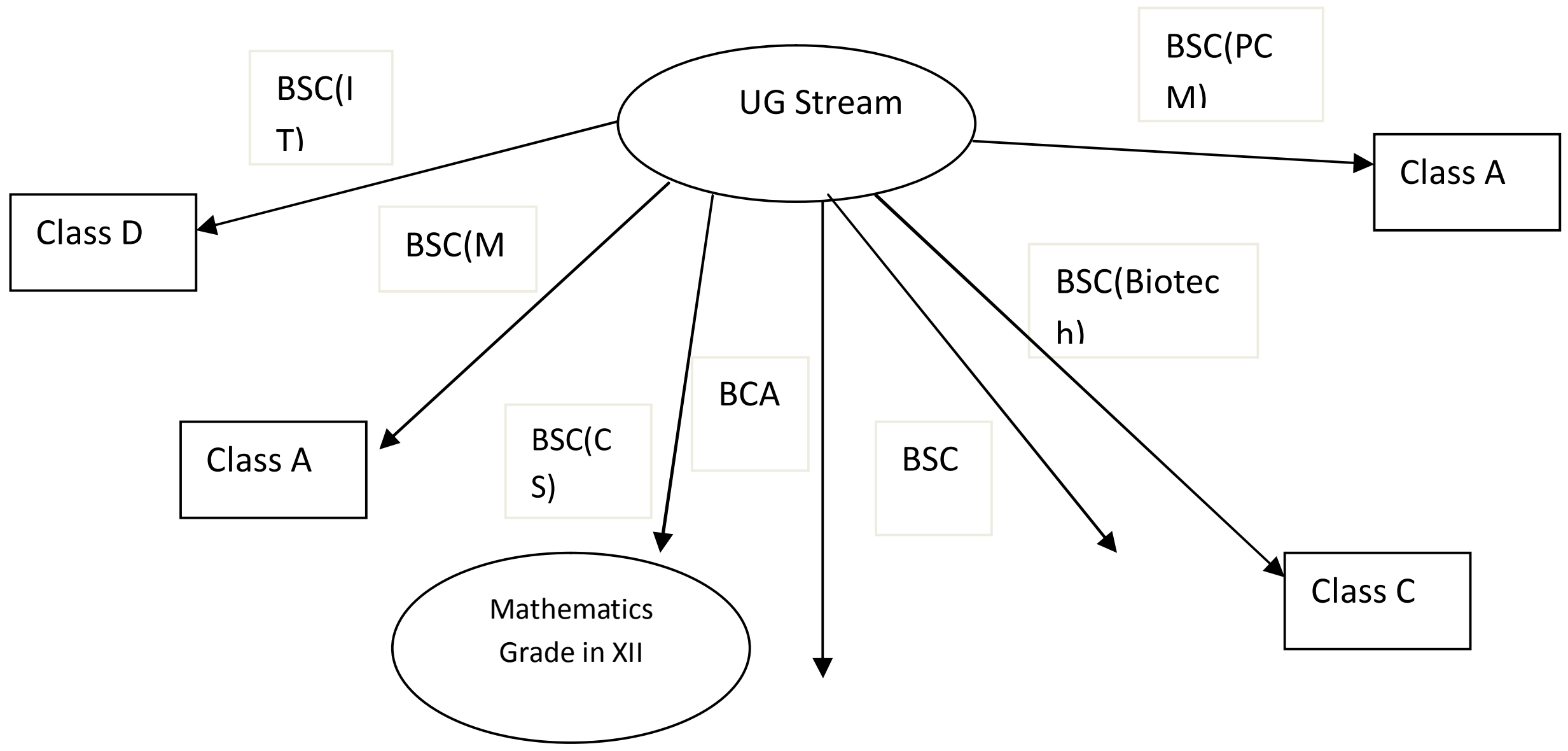

**It's values are A,B,C,D,E,F**

**Since, Entropy (SBSC(CS) MA), Entropy (SBSC(CS)MC), Entropy (SBSC(CS)MD)=0 (No classification required)**

**UG Stream->BSC(CS)->Mathematics Grade in XII->A:B (No further classification)**

**UG Stream->BSC(CS)->Mathematics Grade in XII->C:B (No further classification)**

**UG Stream->BSC(CS)->Mathematics Grade in XII->F:D (No further classification)**

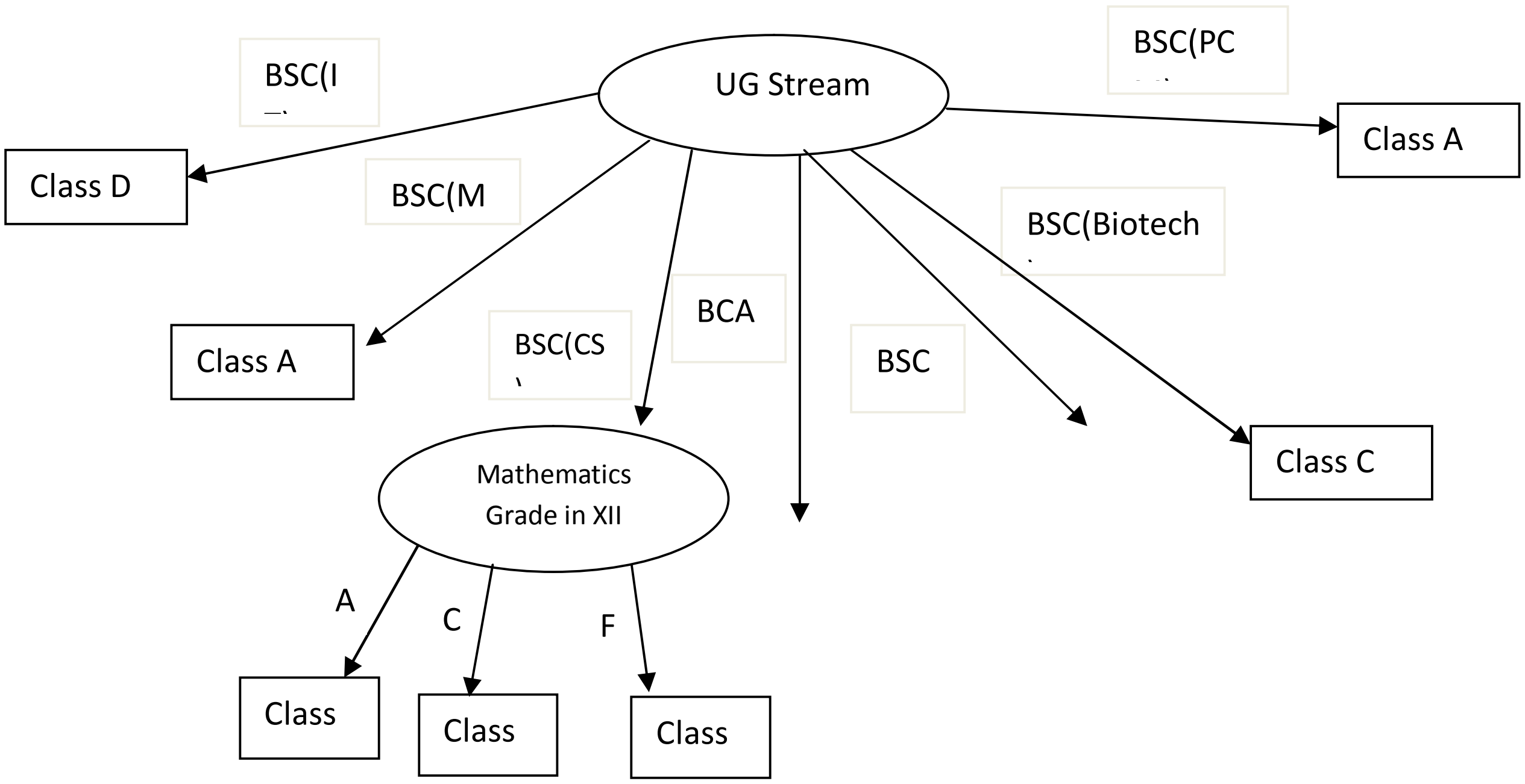

**But Entropy (SBSC(CS)MB)= 0.918296**

**For finding the next node in branch of BSC(CS)->Mathematics Grade in XII->B , calculate the gain value of BSC(CS)MB with other nodes (XII Grade,and UG Grade)**

Gain (SBSC(CS)MB),XII Grade) = ?

XII Grade=A is of occurrence 1(1-B)

XII Grade=B is of occurrence 1(1-B)

XII Grade=C is of occurrence 1(1-D)

Entropy(SBSC(CS)MBXA)=0,Entropy(SBSC(CS)MBXB)=0,
Entropy(SBSC(CSMB)XC)=0

Gain(SBSC(CS)MB),XII Grade)

=Entropy(SBSC(CS)MB)-(1/3)*Entropy(**SBSC(CS)XA)-(1/3)*Entropy(SBSC(CS)XB)-(1/3)* Entropy (SBSC(CS)XC)**

Gain(SBSC(CS)MB),XII Grade) =0.918296-(1/3)*0**-(1/3)*0-(1/3)* 0=0.918296**

Gain (SBSC (CS) MB),UG Grade) = ?

UG Grade =A,D are of occurrence 0

UG Grade =B is of occurrence 2(2-B)

UG Grade =C is of occurrence 1(1-D)

**Entropy (SBSC(CS)MBUA),**

**Entropy (SBSC(CS)MBUB),Entropy (SBSC(CS)MBUC)=** 0,

**Entropy (SBSC(CS)MBUD)=** 0

**Gain(SBSC(CS)MB),UGGrade)**

=Entropy(SBSC(CS)MB)-(0/3)*Entropy(SBSC(CS)MBUA)-(2/3)* Entropy (SBSC(CS)MBUB)-(0/3)* Entropy (SBSC(CS)MBUC)-(0/3)* Entropy (SBSC(CS)MBUD)

Gain(SBSC(CS)MB),UG Grade) =0.918296

Gain(SBSC(CS)MB),XII Grade) & Gain(SBSC(CS)MB),UG Grade) are equal =**0.918296**

**But gain (S,XII Grade) is greater than gain(S,UG Grade) so next node is XII grade in branch**

**UG Stream->BSC(CS)->Mathematics Grade in XII->B->XII Grade**

**XII Grade values are A, B , C, D**

Entropy(SBSC(CS)MBXA)=0,

Entropy(SBSC(CS)MBXB)=0,Entropy(SBSC(CS)MBXC)=0 so there is no classification

**UG Stream->BSC(CS)->Mathematics Grade in XII->B->XII Grade->A:B (No further classification)**

**UG Stream->BSC(CS)->Mathematics Grade in XII->B->XII Grade->B:B (No further classification)**

**UG Stream->BSC(CS)->Mathematics Grade in XII->B->XII Grade->C:D (No further classification).**

**So stop the process here**

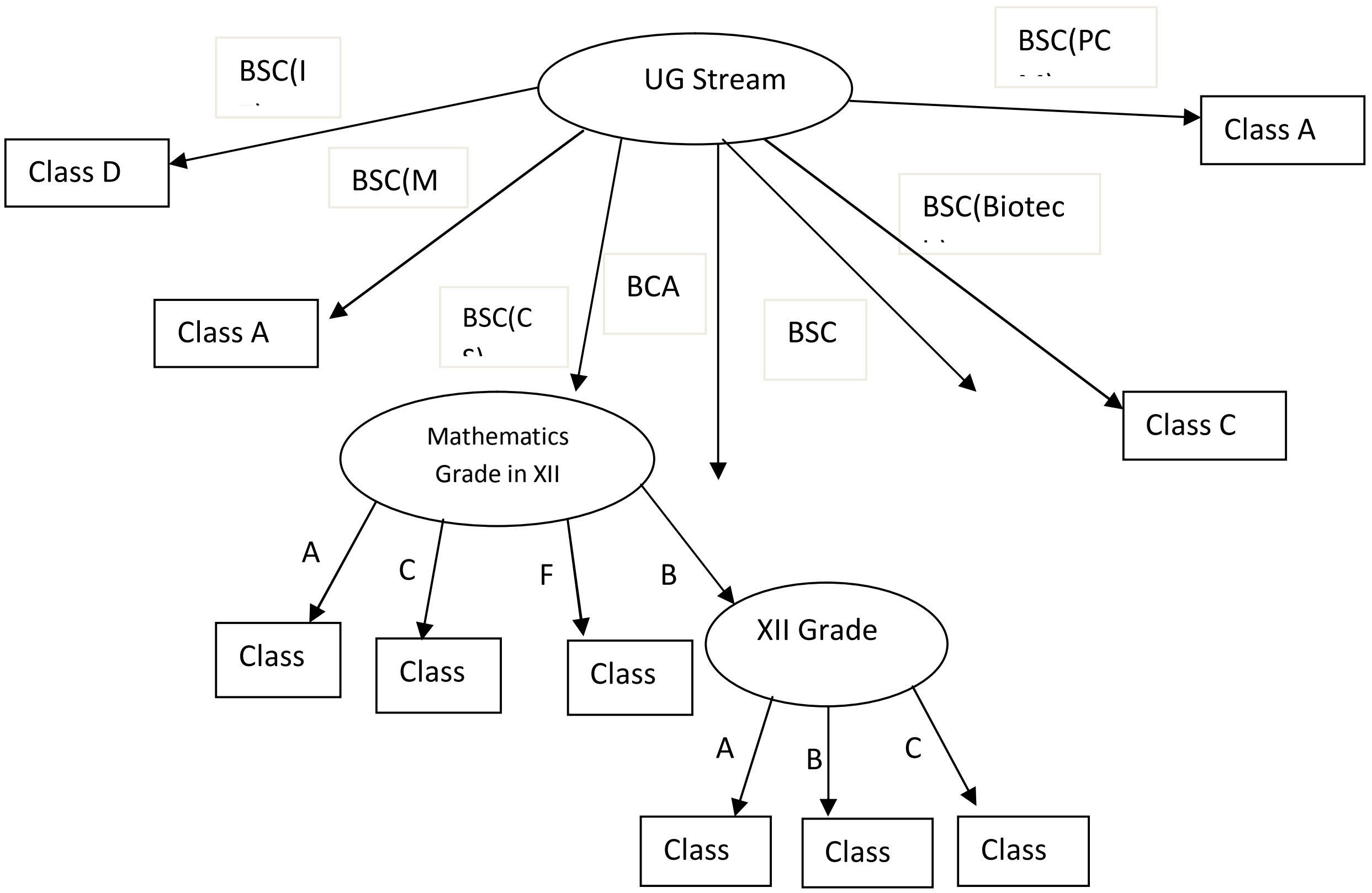

**With UG Stream=BCA**

**2 .For finding the next node in branch of BCA , calculate the gain value of BCA with other nodes (XII Grade, Mathematics Grade in XII and UG Grade)**

a) Gain (SBCA,XII Grade) = ?

XII Grade=A is of occurrence 3( 2-C, 1-D)

XII Grade=B is of occurrence 19( 6-A, 10-B,3-C)

XII Grade=C is of occurrence 7( 1-A, 3-B,3-C)

XII Grade=D is of occurrence 8( 4-C, 4-D)

XII Grade=E is of occurrence 1( 1-C)

**Entropy (SBCAXA)=** -(2/3)xlog2(2/3)-(1/3)xlog2(1/3)= 0.918296

**Entropy (SBCAXB)=** -(6/19)xlog2(6/19)-(10/19)xlog2(10/19) -(3/19)xlog2(3/19)= 1.432983

**Entropy (SBCAXC)=** -(1/7)xlog2(1/7)-(3/7)xlog2(3/7) -(3/7)xlog2(3/7)= 1.08263751

**Entropy (SBCAXD)=** -(4/8)xlog2(4/8)-(4/8)xlog2(4/8)**=1**

**Entropy (SBCAXE)=** -(1/1)xlog2(1/1)=0

**Gain(SBCA,XII Grade) =**Entropy(SBCA)-(3/38)*Entropy(SBCAXA)-(19/38)* Entropy (SBCAXB)-(7/38)* Entropy (SBCAXC)-(8/38)* Entropy (SBCAXD)- (1/38)* Entropy (SBCAXE)

**Gain(SBCA,XII Grade)** =1.893387-(3/38)* 0.918296-(19/38)* 1.432983-(7/38)* 1.082637-(8/38)*1- (1/38)* 0=0.694439

b) Gain ( SBCA, Mathematics Grade in XII) = ?

Mathematics Grade values are A,B,C,D,E and F

Mathematics Grade in XII =A is of occurrence 1( 1-B)

Mathematics Grade in XII =B is of occurrence 17( 6-A,9-B,2-C)

Mathematics Grade in XII =C is of occurrence 6( 1-A,3-B,1-C,1-D)

Mathematics Grade in XII =D is of occurrence 4(1-C,3-D)

Mathematics Grade in XII =F is of occurrence 10( 9-C,1-D)

**Entropy (SBCAMA)=** -(1/1)xlog2(1/1)=0

**Entropy (SBCAMB)=** -(6/17)xlog2(6/17)-(9/17)xlog2(9/17) -(2/17)xlog2(2/17)= 1.37928049

**Entropy (SBCAMC)=** -(1/6)xlog2(1/6)-(3/6)xlog2(3/6) -(1/6)xlog2(1/6) -(1/6)xlog2(1/6)= 1.79248125

**Entropy (SBCAMD)=** -(1/4)xlog2(1/4) -(3/4)xlog2(3/4)= 0.81127812

**Entropy (SBCAMF)=** -(9/10)xlog2(9/10) -(1/10)xlog2(1/10)= 0.46899559

**Gain(SBCA, Mathematics Grade in XII)** =Entropy(SBCA)-(1/38)*Entropy(SBCAMA)-(17/38)* Entropy (SBCAMB)-(6/38)* Entropy (SBCAMC)-(4/38)* Entropy (SBCAMD)- (10/38)* Entropy (SBCAMF)

**Gain(SBCA, Mathematics Grade in XII) =**1.893387-(1/38)*0-(17/38)*1.379280-(6/38)* 1.792481-(4/38)* 0.811278- (10/38)* 0.468995=0.78449995

c) Gain( SBCA, UG Grade) = ?

UG Grade =A is of occurrence 1(1-C)

UG Grade =B is of occurrence 23(7-A,12-B,4-C)

UG Grade =C is of occurrence 11(1-B,7-C,3-D)

UG Grade =D is of occurrence 3(1-C,2-D)

**Entropy (SBCAUA)=** -(1/1)xlog2(1/1)=0

**Entropy (SBCAUB)=** -(7/23)xlog2(7/23) -(12/23)xlog2(12/23) -(4/23)xlog2(4/23)= 1.45090828

**Entropy (SBCAUC)=** -(1/11)xlog2(1/11) -(7/11)xlog2(7/11) -(3/11)xlog2(3/11)= 1.24067053

**Entropy (SBCAUD)=** -(1/3)xlog2(1/3) -(2/3)xlog2(2/3)= 0.91829583

**Gain(SBCA,UG Grade) =**Entropy(SBCA)-(1/38)*Entropy(SBCAUA)-(23/38)* Entropy (SBCAUB)-(11/38)* Entropy (SBCAUC)-(3/38)* Entropy (SBCAUD)

**Gain(SBCA,UG Grade)** =1.893387-(1/38)*0-(23/38)* 1.450908-(11/38)* 1.240670-(3/38)* 0.918295=0.58356755

Gain( SBCA, Mathematics Grade in XII) is greater than Gain(SBCA,XII Grade) and Gain( SBCA, UG Grade) **) so next node is Mathematics grade in XII in branch**

**UG Stream->BCA->Mathematics Grade in XII**

**Mathematics Grade in XII values are A,B,C,D,E and F**

**Since Entropy (SBCAMA)=**0 no classification

**UG Stream->BCA->Mathematics Grade in XII->A: B (**no further classification)

**Entropy (SBCAMB)=** 1.37928049

**Entropy (SBCAMC)=** 1.79248125

**Entropy (SBCAMD)=** 0.81127812

**Entropy (SBCAMF)=** 0.46899559

1) For SBCAMB (**UG Stream->BCA->Mathematics Grade in XII->B :?)**

**For finding the next node in branch UG Stream->BCA->Mathematics Grade in XII->B:? , Calculate the gain value of SBCAMB with other nodes (XII Grade, UG Grade).**

a) Gain (SBCAMB, XII Grade) = ?

XII Grade=A is of occurrence 0

XII Grade=B is of occurrence 17( 6-A,10-B,1-C)

XII Grade=C is of occurrence 0

XII Grade=D is of occurrence

**Entropy (SBCAMBXB)=** -(6/17)xlog2(6/17) -(10/17)xlog2(10/17) -(1/17)xlog2(1/17)= 1.221047

**Gain(SBCAMB,XII Grade) =**Entropy(SBCAMB)-(17/17)*Entropy(SBCAMBXB)

**Gain(SBCAMB,XII Grade) =**1.37928049-(17/17)* 1.221047=0.15823349

b) Gain (SBCAMB, UG Grade) = ?

UG Grade=A is of occurrence 1(1-C)

UG Grade=B is of occurrence 15(6-A,8-B,1-C)

UG Grade=C is of occurrence 1 (1-B)

UG Grade=D is of occurrence 0(null)

UG Grade=E is of occurrence 0(null)

UG Grade=F is of occurrence 0(null)

**Entropy (SBCAMBUA)=** -(1/1)xlog2(1/1) =0

**Entropy (SBCAMBUB)=** -(6/15)xlog2(6/15) -(8/15)xlog2(8/15) -(1/15)xlog2(1/15)= 1.272905

**Entropy (SBCAMBUC)=** -(1/1)xlog2(1/1) =0

**Gain(SBCAMB,UG Grade) =** Entropy(SBCAMB )-(15/17)* Entropy (SBCAMBUB)

**Gain(SBCAMB,UG Grade) =**1.379280-(15/17)* 1.272905=0.25612853

Gain(SBCAMB,UG Grade) is greater than Gain(SBCAMB,XII Grade) so UG Grade is next node in branch **UG Stream->BCA->Mathematics Grade in XII->B->UG Grade**

**For finding the next node in branch UG Stream->BCA->Mathematics Grade in XII->B->UG Grade->? , Calculate the gain value of it's with other nodes (XII Grade).**

**UG Grade values are A, B, C, D and E**

**Entropy (SBCAMBUA)**, **Entropy (SBCAMBUC)** =0

**Entropy (SBCAMBUB) =** 1.272905

Gain (SBCAMBUB, XII Grade) =?

XII Grade values are A, B, C, D, E

XII Grade=A,C,D,E,F are of occurrence 0

XII Grade=B is of occurrence 15(6-A8-B, 1-C)

Entropy(SBCAMBUBXB)= 1.272905

**Gain(SBCAMBUB,XII Grade) =** Entropy (SBCAMBUB)-(15/15)* Entropy (SBCAMBUBXB)

**Gain(SBCAMBUB,XII Grade) =**1.272905-(15/15)* 1.272905=0 (no classification)

2) For SBCAMC (**UG Stream->BCA->Mathematics Grade in XII->C :?)**

**For finding the next node in branch UG Stream->BCA->Mathematics Grade in XII->C:? , Calculate the gain value of SBCAMC with other nodes (XII Grade, UG Grade).**

a) Gain (SBCAMC, XII Grade) = ?

XII Grade=A,B are of occurrence 0

XII Grade=C is of occurrence 5(1-A,3-B,1-C)

XII Grade=D is of occurrence 1(1-D)

Entropy(SBCAMCXC)= -(1/5)xlog2(1/5) -(3/5)xlog2(3/5) -(1/5)xlog2(1/5)= 1.370951

Entropy(SBCAMCXD)= -(1/1)xlog2(1/1)=0

**Gain(SBCAMC,XIIGrade)=**Entropy(SBCAMC)-(5/6)*Entropy(SBCAMCXC)-(1/6)* Entropy(SBCAMCXD)

**Gain(SBCAMC,XII Grade)=** 1.79248125 -(5/6)* 1.370951-(1/6)* **0=**0.650022

b) Gain (SBCAMC, UG Grade) = ?

UG Grade values are A,B,C,D and E

UG Grade=A,D,E are of occurrence 0

UG Grade=B is of occurrence 4(1-A,3-B)

UG Grade=C is of occurrence 2(1-C,1-D)

Entropy(SBCAMCUB)= -(1/4)xlog2(1/4) -(3/4)xlog2(3/4)= 0.811278

Entropy(SBCAMCUC)= -(1/2)xlog2(1/2) -(1/2)xlog2(1/2)=1

**Gain (SBCAMC, UG Grade) =** Entropy (SBCAMC)-(4/6)* Entropy (SBCAMCUB) -(2/6)* Entropy (SBCAMCUC)

**Gain (SBCAMC, UG Grade) =**1.79248125-(4/6)* 0.811278-(2/6)* 1=0.918296

Gain(SBCAMC,UG Grade) is greater than Gain(SBCAMC,XII Grade) so UG Grade is next node in branch

**UG Stream->BCA->Mathematics Grade in XII->C->UG Grade**

**For finding the next node in branch UG Stream->BCA->Mathematics Grade in XII->C->UG Grade->? , Calculate the gain value of it's with other nodes (XII Grade).**

**UG Grade values are A, B, C, D and E**

**Entropy (SBCAMCUA), Entropy (SBCAMCUD), Entropy (SBCAMCUE)** =0

Entropy(SBCAMCUB)=0.811278

Gain(SBCAMCUB,XII Grade)=?

XII Grade values are A, B, C, D, E

XII Grade=A,B are of occurrence 0

XII Grade=C is of occurrence 4(1-A,3-B)

**Entropy(SBCAMCUBXC)=** -(1/4)xlog2(1/4) -(3/4)xlog2(3/4)= 0.811278

**Gain (SBCAMCUB,XII Grade) =** Entropy (SBCAMCUB)-(4/4)* Entropy (SBCAMCUBXC)

**Gain (SBCAMCUB,XII Grade) =**0.811278-(4/4)* 0.811278=0 no classification

**UG Stream->BCA->Mathematics Grade in XII->C->UG Grade->B:B (**no further classification)

Entropy (SBCAMCUC)=1

Gain (SBCAMCUC, XII Grade) =?, XII Grade values are A, B, C, D, E

XII Grade=A,B are of occurrence 0

XII Grade=C is of occurrence 1(1-C)

XII Grade=D is of occurrence 1(1-D)

Entropy(SBCAMCUCXC)= -(1/1)xlog2(1/1)=0,Entropy(SBCAMCUCXD)= -(1/1)xlog2(1/1)=0

**Gain (SBCAMCUC, XII Grade) =** Entropy (SBCAMCUC)-(1/2)* Entropy (SBCAMCUCXC)-(1/2)*Entropy(SBCAMCUCXD)

**Gain (SBCAMCUC, XII Grade) =**1-(1/2)*0-(1/2)*0=1 classification is required

Since Entropy(SBCAMCUCXC) and Entropy(SBCAMCUCXD)= 0 no classification further

**UG Stream->BCA->Mathematics Grade in XII->C->UG Grade->C->XII Grade->C:C (**no further classification)

**UG Stream->BCA->Mathematics Grade in XII->C->UG Grade->C->XII Grade->D:D (**no further classification)

2) For SBCAMD (**UG Stream->BCA->Mathematics Grade in XII->D :?)**

**For finding the next node in branch UG Stream->BCA->Mathematics Grade in XII->D:? , Calculate the gain value of SBCAMD with other nodes (XII Grade, UG Grade).**

a) Gain (SBCAMD, XII Grade) = ?

XII Grade values are A, B, C, D, E

XII Grade=A,C,E are of occurrence 0

XII Grade=B is of occurrence 1(1-C)

XII Grade=D is of occurrence 3(3-D)

Entropy(SBCAMDXB)= -(1/1)xlog2(1/1)=0

Entropy(SBCAMDXD)= -(3/3)xlog2(3/3)=0

**Gain (SBCAMD, XII Grade) =** Entropy (SBCAMD)-(1/4)* Entropy (SBCAMDXB)-(3/4)*Entropy(SBCAMDXD)

**Gain (SBCAMD, XII Grade) =**0.81127812 -(1/4)*0-(3/4)*0=0.811278

b) Gain (SBCAMD, UG Grade) = ?

UG Grade values are A,B,C,D and E

UG Grade=A,B are of occurrence 0

UG Grade=C is of occurrence 2(1-D,1-C)

UG Grade=D is of occurrence 2(2-D)

Entropy(SBCAMDUC)= -(1/2)xlog2(1/2) -(1/2)xlog2(1/2)=1

Entropy(SBCAMDUD)= -(2/2)xlog2(2/2)=0

**Gain (SBCAMD, UG Grade) =** Entropy (SBCAMD)-(2/4)* Entropy (SBCAMDUC)-(2/4)*Entropy(SBCAMDUD)

**Gain (SBCAMD, UG Grade) =** 0.81127812-(2/4)*1-(2/4)*0=0.311278

Gain (SBCAMD, XII Grade) is greater than Gain (SBCAMD, UG Grade) so XII Grade is next node in branch

**UG Stream->BCA->Mathematics Grade in XII->D->XII Grade**

**XII Grade values are A,B,C,D,E**

Entropy(SBCAMDXB)= -(1/1)xlog2(1/1)=0 , Entropy(SBCAMDXD)= -(3/3)xlog2(3/3)=0 **(**no further classification)

**UG Stream->BCA->Mathematics Grade in XII->D-> XII Grade->B:C (**no further classification)

**UG Stream->BCA->Mathematics Grade in XII->D-> XII Grade->D:D (**no further classification)

2) For SBCAMF (**UG Stream->BCA->Mathematics Grade in XII->F :?)**

**For finding the next node in branch UG Stream->BCA->Mathematics Grade in XII->F:? , Calculate the gain value of SBCAMF with other nodes (XII Grade, UG Grade).**

a) Gain (SBCAMF, XII Grade) = ?

XII Grade values are A, B, C, D, E

XII Grade=A is of occurrence 3(2-C,1-D)

XII Grade=B is of occurrence 0

XII Grade=C is of occurrence 2(2-C)

XII Grade=D is of occurrence 4(4-C)

XII Grade=E is of occurrence 1(1-C)

Entropy(SBCAMFXA)= -(2/3)xlog2(2/3) -(1/3)xlog2(1/3)= 0.918296

Entropy(SBCAMFXC)= -(2/2)xlog2(2/2)=0,Entropy(SBCAMFXD)= -(4/4)xlog2(4/4)=0

Entropy(SBCAMFXE)= -(1/1)xlog2(1/1)=0

**Gain (SBCAMF, XII Grade)**

= Entropy (SBCAMF)-(3/10)* Entropy (SBCAMFXA)-(2/10)*Entropy(SBCAMFXC) - (4/10)* Entropy (SBCAMFXD)-(1/10)*Entropy(SBCAMFXE)

**Gain (SBCAMF, XII Grade)** =0.46899559-(3/10)*0.918296-(2/10)*0-(4/10)*0-(1/10)*0=0.193507

b) Gain (SBCAMF, UG Grade) = ?

UG Grade values are A,B,C,D and E

UG Grade=A,E are of occurrence 0

UG Grade=B is of occurrence 3(3-C)

UG Grade=C is of occurrence 6(5-C,1-D)

UG Grade=D is of occurrence 1(1-C)

Entropy(SBCAMFUB)= -(3/3)xlog2(3/3)=0

Entropy(SBCAMFUC)= -(5/6)xlog2(5/6) -(1/6)xlog2(1/6)= 0.650022

Entropy(SBCAMFUD)= -(1/1)xlog2(1/1)=0

**Gain (SBCAMF, UG Grade) =** Entropy (SBCAMF)-(3/10)* Entropy (SBCAMFUB)-(6/10)*Entropy(SBCAMFUC) -(1/10)* Entropy (SBCAMFUD)

**Gain (SBCAMF, UG Grade)** =0.46899559-(3/10)*0-(6/10)* 0.650022 -(1/10)* 0= 0.078982

Gain (SBCAMF, XII Grade) is greater than Gain (SBCAMF, UG Grade) so XII grade is next node in branch

**UG Stream->BCA->Mathematics Grade in XII->F-> XII Grade**

**For finding the next node in branch UG Stream->BCA->Mathematics Grade in XII->F->XII Grade->? , Calculate the gain value of it's with other nodes (UG Grade).**

**XII Grade values are A,B,C,D,E**

Entropy(SBCAMFXA)= 0.918296

Entropy(SBCAMFXC),Entropy(SBCAMFXD),Entropy(SBCAMFXE)= 0

(No further classification )

Gain(SBCAMFXA,UG Grade)=?

UG Grade values are A,B,C,D and E

UG Grade=A,D,E are of occurrence 0 (null)

UG Grade=B is of occurrence 1(1-C)

UG Grade=C is of occurrence 2(1-C, 1-D)

Entropy(SBCAMFXAUB)= -(1/1)xlog2(1/1)=0

Entropy(SBCAMFXAUC)= -(1/2)xlog2(1/2) -(1/2)xlog2(1/2)=1

**Gain(SBCAMFXA,UGGrade)=**

Entropy(SBCAMFXA)-(1/3)*Entropy(SBCAMFXAUB)-(2/3)*Entropy(SBCAMFXAUC)

**Gain(SBCAMFXA,UG Grade)=** 0.918296-(1/3)*0-(2/3)*1= 0.251629

Since Entropy(SBCAMFXAUB)= 0 (no further classification)

**UG Stream->BCA->Mathematics Grade in XII->F-> XII Grade->A->UG Grade->B:C (**no further classification)

Entropy(SBCAMFXAUC)= 1 there is classification required but there is no remaining attribute so stop the process

**UG Stream->BCA->Mathematics Grade in XII->F-> XII Grade->A->UG Grade->C:D (**no further classification)

With UG Stream=BSC

3. For finding the next node in branch of BSC , calculate the gain value of BSC with other nodes (XII Grade, Mathematics Grade in XII and UG Grade)

a) Gain (SBSC, XII Grade) = ?

XII Grade values are A, B, C, D, E

XII Grade=A is of occurrence 1(1-C)

XII Grade=B is of occurrence 1(1-C)

XII Grade=C is of occurrence 6 (2-C,4-D)

XII Grade=D is of occurrence 2(1-C,1-D)

XII Grade=E is of occurrence 0 (null)

Entropy(SBSCXA)= -(1/1)xlog2(1/1)=0,Entropy(SBSCXB)= -(1/1)xlog2(1/1)=0

Entropy(SBSCXC)= -(2/6)xlog2(2/6) -(4/6)xlog2(4/6)= 0.918296

Entropy(SBSCXD)= -(1/2)xlog2(1/2) -(1/2)xlog2(1/2)=1

Gain(SBSC,XIIGrade)=
Entropy(SBSC)-(1/10)*Entropy(SBSCXA)-(1/10)*Entropy(SBSCXB)-(6/10)*Entropy (SBSCXC)-(2/10)*Entropy(SBSCXD)

Gain(SBSC,XIIGrade)=1-(1/10)*0-(1/10)*0-(6/10)* 0.918296-(2/10)*1=0.249022

b) Gain (SBSC, Mathematics Grade in XII) = ?

Mathematics Grade values are A,B,C,D,E and F

Mathematics Grade in XII =A,D,E are of occurrence 0 (null)

Mathematics Grade in XII =B is of occurrence 1( 1-C)

Mathematics Grade in XII =C is of occurrence 4( 1-C,3-D)

Mathematics Grade in XII =F is of occurrence 5( 3-C,2-D)

Entropy(SBSCMB)= -(1/1)xlog2(1/1)=0

Entropy(SBSCMC)= -(1/4)xlog2(1/4)-(3/4)xlog2(3/4)= 0.811278

Entropy(SBSCMF)= -(3/5)xlog2(3/5)-(2/5)xlog2(2/5)= 0.970951

**Gain (SBSC, Mathematics Grade in XII) =**

Entropy(SBSC)-(1/10)*Entropy(SBSCMB)-(4/10)*Entropy(SBSCMC)-(5/10)*Entropy(SBSCMF)

**Gain (SBSC, Mathematics Grade in XII)** =1-(1/10)*0-(4/10)* 0.811278-(5/10)* 0.970951=0.190013

c) Gain (SBSC,UG Grade) = ?

UG Grade values are A,B,C,D and E

UG Grade=A,B,D,E are of occurrence 0 (null)

UG Grade=C is of occurrence 10 (5-C,5-D)

Entropy(SBSCUC)= -(5/10)xlog2(5/10)-(5/10)xlog2(5/10)=1

Gain (SBSC,UG Grade) = Entropy (SBSC)-(10/10)* Entropy (SBSCUC)

Gain (SBSC,UG Grade) =1-(10/10)*1=0

Gain(SBSC,XII Grade) is greater than Gain (SBSC, Mathematics Grade in XII) & Gain(SBSC,UG Grade) so XII Grade is next node in branch :- **UG Stream->BSC->XII Grade**

For finding the next node in branch of **UG Stream->BSC->XII Grade->?** , calculate the gain value of it's with other nodes (Mathematics Grade in XII and UG Grade)

XII Grade values are A, B, C, D, E

Entropy (SBSCXA) and Entropy (SBSCXB)= 0( No further classification)

Entropy (SBSCXC)= 0.918296

Entropy (SBSCXD)= 1

**1) For SBSCXC**

**For finding the next node in branch of UG Stream->BSC->XII Grade->C->? , calculate the gain value of it's with other nodes (Mathematics Grade in XII and UG Grade)**

a) Gain(SBSCXC, Mathematics Grade in XII)=?

Mathematics Grade values are A,B,C,D,E and F

Mathematics Grade in XII =A,B,D,E are of occurrence 0 (null)

Mathematics Grade in XII =C is of occurrence 3 (3-D)

Mathematics Grade in XII =F is of occurrence 3 (2-C,1-D)

Entropy(SBSCXCMC)= -(3/6)xlog2(3/6)= 0.5

Entropy(SBSCXCMF)= -(2/3)xlog2(2/3)-(1/3)xlog2(1/3) = 0.918296

**Gain(SBSCXC, Mathematics Grade in XII)=**

Entropy(SBSCXC)-(3/6)* Entropy(SBSCXCMC)-(3/6)*Entropy(SBSCXCMF)

**Gain(SBSCXC, Mathematics Grade in XII)=**

0.918296-(3/6)* 0.5-(3/6)* 0.918296=0.209148

**b) Gain(SBSCXC,UG Grade)=?**

**UG Grade values are A,B,C,D and E**

UG Grade=A,B,D,E are of occurrence 0 (null)

UG Grade=C is of occurrence 6 (2-C,4-D)

Entropy(SBSCXCUC)= -(2/6)xlog2(2/6)-(4/6)xlog2(4/6)= 0.918296

**Gain(SBSCXC,UG Grade)=** Entropy(SBSCXC)-(6/6)* Entropy(SBSCXCUC)

**Gain(SBSCXC,UG Grade)=** 0.918296-(6/6)* 0.918296=0

Gain(SBSCXC, Mathematics Grade in XII) is highest than Gain(SBSCXC,UG Grade) so Mathematics Grade in XII is next node in branch

**UG Stream->BSC->XII Grade->C->Mathematics Grade in XII**

For finding the next node in branch of **UG Stream->BSC->XII Grade->C->Mathematics Grade in XII->?**, calculate the gain value of it's with other node (UG Grade)

Mathematics Grade values are A,B,C,D,E and F

Mathematics Grade in XII =A,B,D,E are of occurrence 0 (null)

Entropy(SBSCXCMC)= 0.5

Entropy(SBSCXCMF)= 0.918296

**<u>For SBSCXCMC</u>**

**For finding the next node in branch of UG Stream->BSC->XII Grade->C->Mathematics Grade in XII->C->?, calculate the gain value of it's with other node (UG Grade)**

Gain(SBSCXCMC,UG Grade)= ?

UG Grade values are A,B,C,D and E

UG Grade=A,B,D,E are of occurrence 0 (null)

UG Grade=C is of occurrence 3 (3-D)

Entropy(SBSCXCMCUC)= -(3/3)xlog2(3/3)=0 (No further classification)

**Gain(SBSCXCMC,UG Grade)=**

Entropy(SBSCXCMC)-(3/3)* Entropy(SBSCXCMCUC)

**Gain(SBSCXCMC,UG Grade)=** 0.5-(3/3)*0=0.5

Since Entropy(SBSCXCMCUC)= 0 (no further classification)

**UG Stream->BSC->XII Grade->C->Mathematics Grade in XII->C: D (**no further classification)

**For SBSCXCMF**

**For finding the next node in branch of UG Stream->BSC->XII Grade->C->Mathematics Grade in XII->F->?, calculate the gain value of it's with other node (UG Grade)**

Gain(SBSCXCMF,UG Grade)= ?

UG Grade values are A,B,C,D and E

UG Grade=A,B,D,E are of occurrence 0 (null)

UG Grade=C is of occurrence 3 (2-C,1-D)

Entropy(SBSCXCMFUC)= -(2/3)xlog2(2/3) -(1/3)xlog2(1/3)= 0.918296

**Gain(SBSCXCMF,UG Grade)=**

Entropy(SBSCXCMF)-(3/3)* Entropy(SBSCXCMFUC)

**Gain(SBSCXCMF,UG Grade)=**

0.918296-(3/3)* 0.918296=0 (No further classification)

**UG Stream->BSC->XII Grade->C->Mathematics Grade in XII->F: C (**no further classification)

**2) For SBSCXD**

**For finding the next node in branch of UG Stream->BSC->XII Grade->D->? , calculate the gain value of it's with other nodes (Mathematics Grade in XII and UG Grade)**

Since Entropy(SBSCXD)= 1

a) Gain(SBSCXD, Mathematics Grade in XII)=?

Mathematics Grade values are A,B,C,D,E and F

Mathematics Grade in XII =A,B,D,E are of occurrence 0 (null)

Mathematics Grade in XII =C is of occurrence 1 (1-C)

Mathematics Grade in XII =F is of occurrence 1 (1-D)

Entropy(SBSCXDMC)= -(1/1)xlog2(1/1)=0,Entropy(SBSCXDMF)= -(1/1)xlog2(1/1)=0 (No further classification)

Gain(SBSCXD, Mathematics Grade in XII)=

Entropy(SBSCXD)-(1/2)* Entropy(SBSCXDMC)-(1/2)* Entropy(SBSCXDMF)

Gain(SBSCXD, Mathematics Grade in XII)= 1-(1/2)*0-(1/2)* 0=1

b) Gain(SBSCXD,UG Grade)=?

UG Grade values are A,B,C,D and E

UG Grade=A,B,D,E are of occurrence 0 (null)

UG Grade=C is of occurrence 2 (1-C,1-D)

Entropy(SBSCXDUC)= -(1/2)xlog2(1/2) -(1/2)xlog2(1/2)=1

Gain(SBSCXD,UG Grade)= Entropy(SBSCXD)-(2/2)* Entropy(SBSCXDUC)

Gain(SBSCXD,UG Grade)= 1-(2/2)* 1=0

Gain(SBSCXD, Mathematics Grade in XII) is higher than Gain(SBSCXD,UG Grade) so Mathematics Grade is next node in branch **UG Stream->BSC->XII Grade->D->Mathematics Grade in XII**

**For finding the next node in branch of UG Stream->BSC->XII Grade->D->Mathematics Grade in XII->? , calculate the gain value of it's with other node (UG Grade)**

Mathematics Grade in XII =A,B,D,E are of occurrence 0 (null)

Entropy(SBSCXDMC),Entropy(SBSCXDMF)= 0 (No further classification)

**UG Stream->BSC->XII Grade->D->Mathematics Grade in XII->C: C (No further classification)**

**UG Stream->BSC->XII Grade->D->Mathematics Grade in XII->F: D (No further classification)**

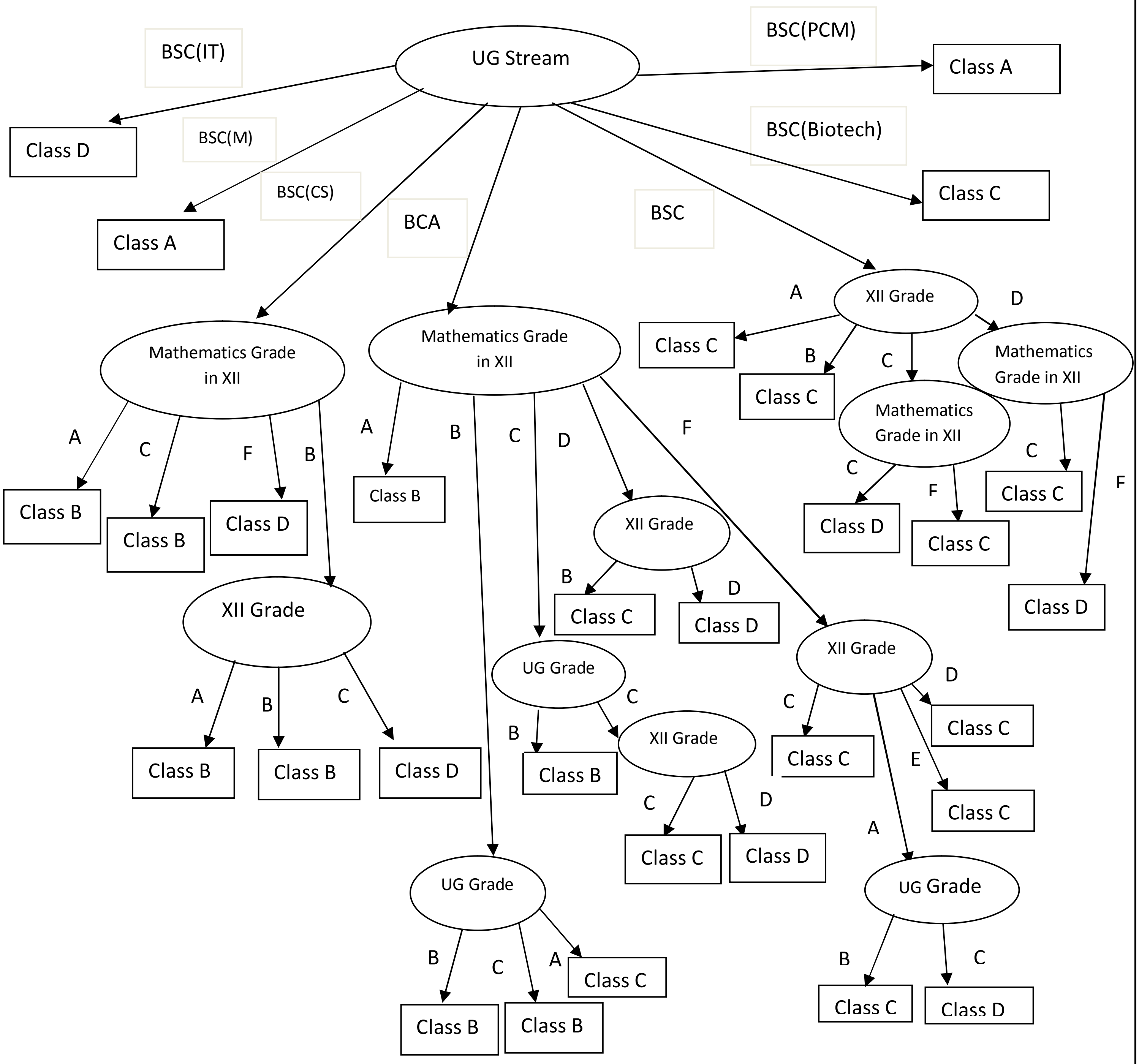

**Figure 3.16: Final generated ID3 Decision tree for MCA student performance**

## 3.1.5 Conclusion

In this study, a reasonably accurate model has been built that helps academicians and administrators to predict student's enrollment in MCA course. This study examines that student's performance(past academic)can be used to construct a model using classification with a decision tree algorithm (ID3 and J48 decision tree algorithm).The concept of classification with Decision Tree helps to extracting the knowledge from the past academic data. In the confusion matrix of ID3 and J48, it is shown that out of four actual categories, the accuracy of D class is 78.3% and 69.6% respectively that means this model is successfully identifying the which student doesn't perform better in MCA. So, there is a need for proper counseling for the students that are going to select in MCA course. This study also helps students in selecting the course for admission according to his skills and academics. It results in that B.Sc. students with Mathematics and BCA stream students performed better in MCA but B.Sc. without Mathematics stream students did not perform well.

## 3.1.6 Testing

We can make predictions for a test set, whether that set contains valid class values or not. The output will contain both the actual and predicted class.

**Classifying new data**

new instance:- **{B, A, BSC(Math), ?}.**

(The class attribute is ? because you don't know the classification.)

Do the following:

STEP1: Guess a value for **?** (or set it at random), say **B**, i.e.

**{B, A, BSC(Math), B}**

STEP 2: Then create a test file and include the above instance .

STEP 3: Click on the **Choose** button and choose a classifier (trees->ID3 decision tree algorithm).

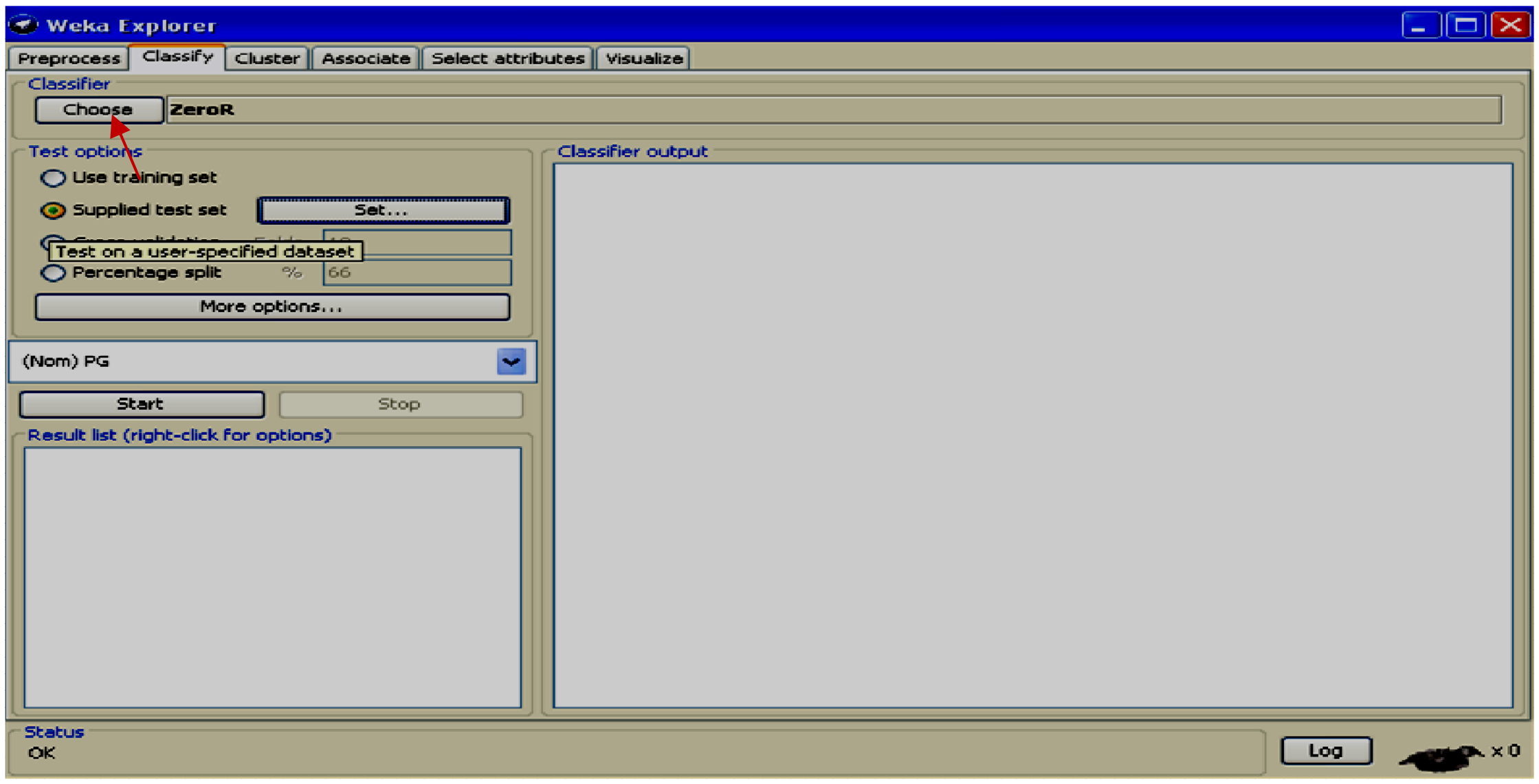

**Figure 3.17 : selecting a classifier from WEKA Explorer**

STEP 4: open the test set file using supplied test set

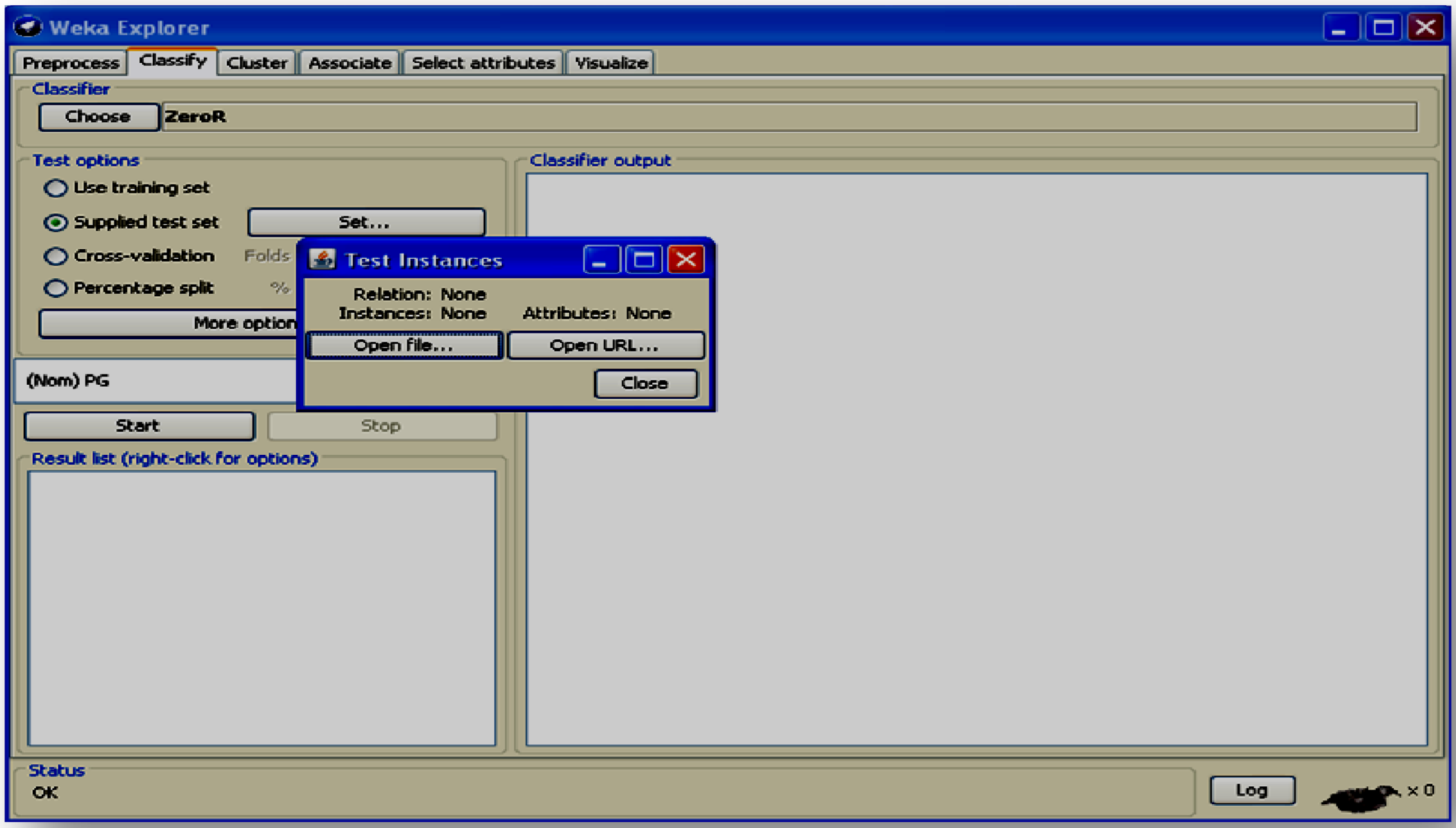

**Figure 3.18:Open the test set file**

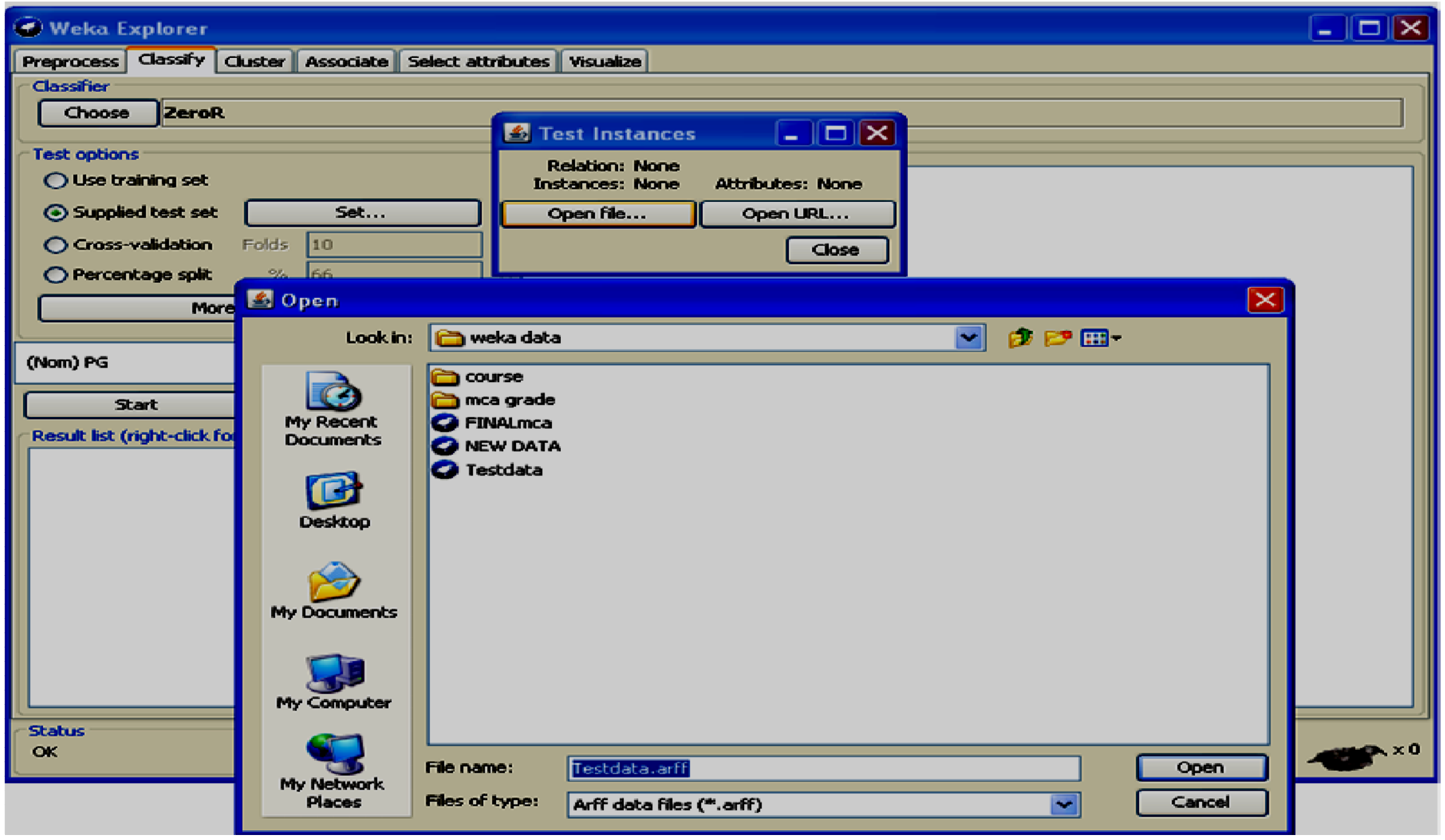

STEP 5: Select more options for classifier evaluation

I. Click on More options

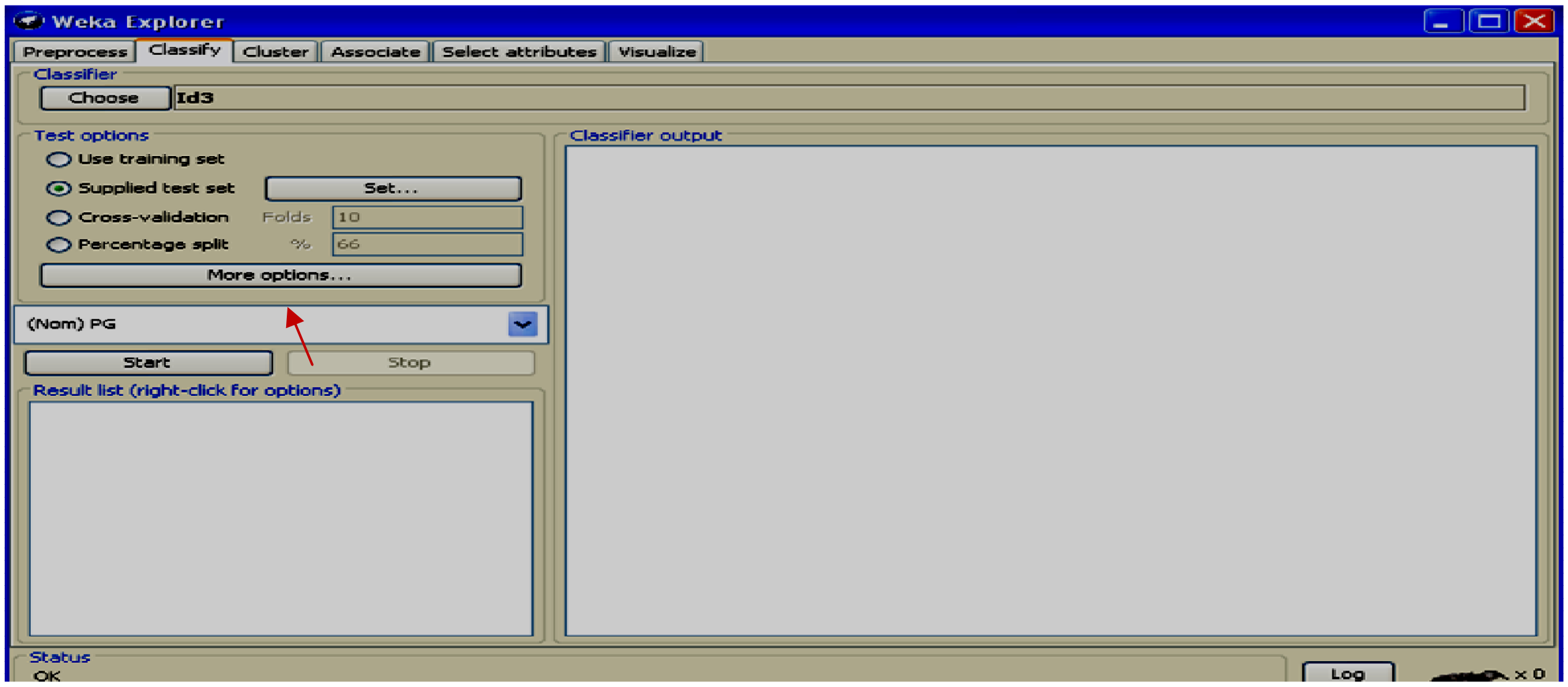

**Figure 3.19: Selecting more options**

II. Click on some evaluation options as selected in figure

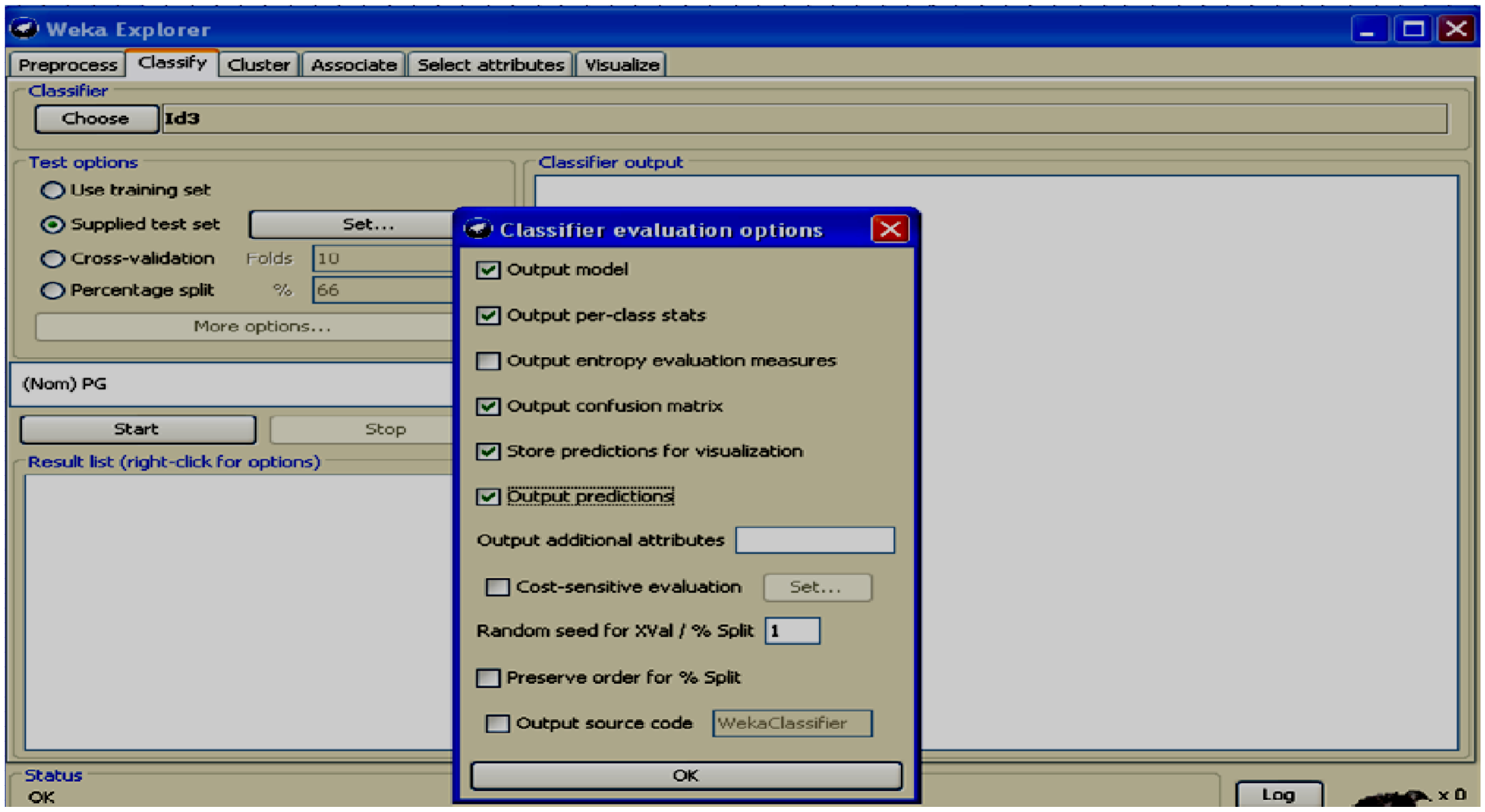

STEP 6: Start the classification

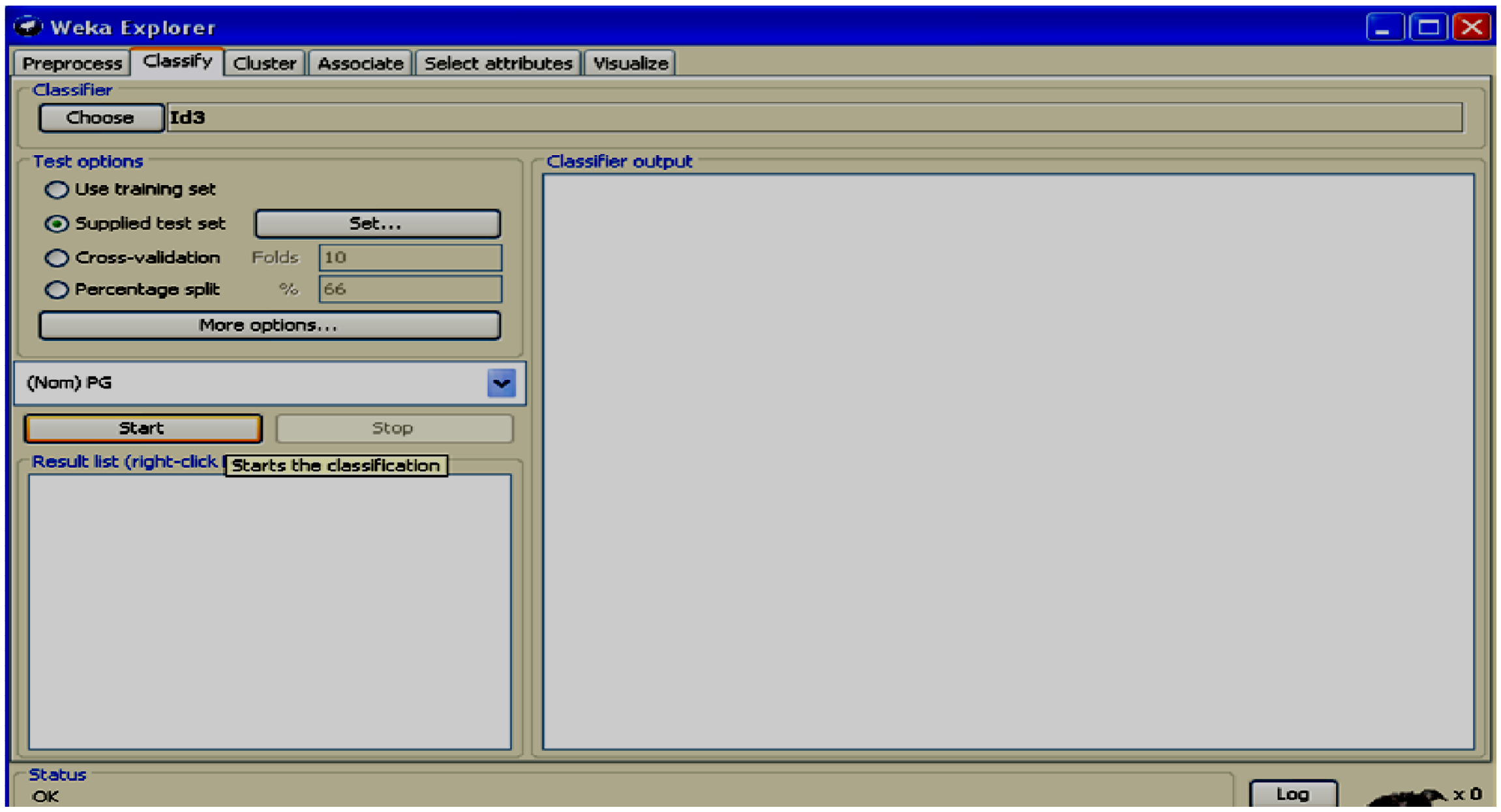

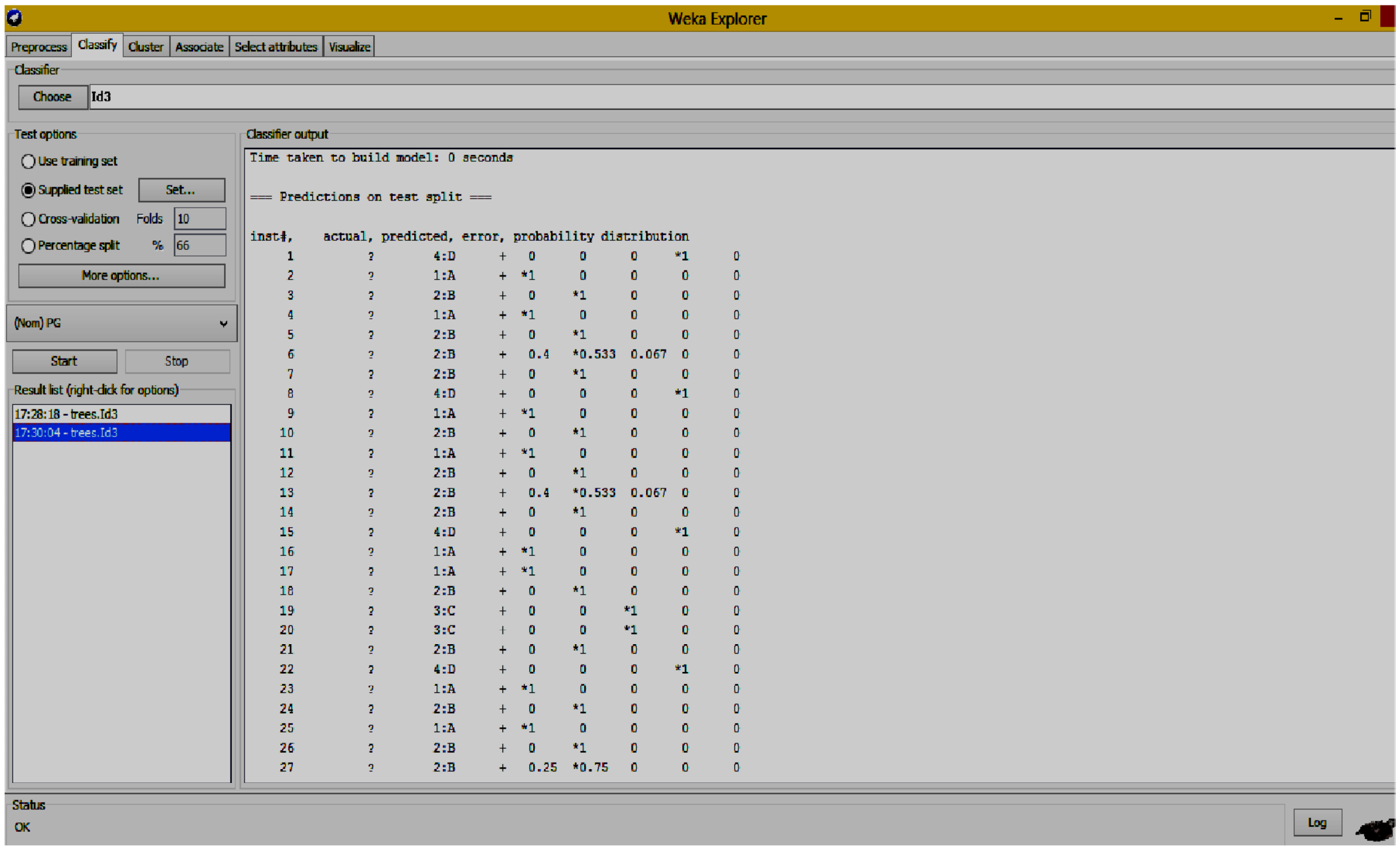

**Figure 3.20: Test set prediction with unknown actual classification**

## 3.2 DATA MINING APPLICATION IN ADVERTISEMENT MANAGEMENT OF HIGHER EDUCATIONAL INSTITUTES

In recent years, Indian higher educational institute's competition grows rapidly for attracting students to get enrollment in their institutes. To attract students educational institutes select a best advertisement method. There are different advertisements available in the market but a selection of them is very difficult for institutes. This study is helpful for institutes to select a best advertisement medium using some data mining methods.

### 3.2.1 Data Used

In this study the data is collected through the questionnaire survey at the institute. There are 500 questionnaires are collected. This questionnaire includes student personnel and academic information. This study is helpful for finding a better advertisement for institutes so this questionnaire includes a important question "How They knows about this university"?

Table 3.10 shows different advertisement methods that are available in questionnaire's field. Table 3.11 shows different combination of advertisements occurrences that occur while responding the questionnaire field.

**Table 3.10: Different advertisement methods**

| **Advertising methods** | **Codes** | **Answers** |
|---|---|---|
| **Friends** | **FR** | **309** |
| Family | FA | 17 |
| Internet Search | IS | 24 |
| Online Advertising | OA | 8 |
| Newspaper | NE | 8 |
| Others | OT | 5 |

**Table 3.11: Different combination of advertisements**

| **Relations** | **Occurrences** |
|---|---|
| Friends and Family(FRFA) | 10 |
| Friends and Internet Search (FRIS) | 6 |
| Friends and Online Advertising(FROA) | 2 |
| Friends and Newspaper (FRNE) | 3 |
| Friends and Others(FROT) | 0 |
| Family and Internet Search(FAIS) | 0 |
| Family and Online Advertising(FAOA) | 0 |
| Family and Newspaper (FANE) | 7 |
| Family and Others(FAOT) | 0 |
| Internet Search and Online Advertising(ISOA) | 4 |
| Internet Search and Newspaper(ISNE) | 2 |
| Internet Search and Others(ISOT) | 0 |
| Online Advertising and Newspaper(OANE) | 2 |
| Online Advertising | 0 |
| and Others(OAOT) | |
| Newspaper and Others(NEOT) | 0 |
| **Total counts(support)** | **36** |

## 3.2.2 Analysis of Support and Confidence Value

The support value of a data item indicates the percentage of database transaction in which that data item appears. Confidence value measures the rule's strength. A small support and large confidence value are meaningful. Support and confidence analysis is shown in Table 3.12.Support and Confidence values are calculated as:

**Support= Occurrences/Total Support**
**Confidence=Support(X,Y)/Support(X)**

**Table 3.12: Support and confidence analysis of different relations**

| **Relations** | **Support = occurrences/Total Support** | **Confidence = support(X,Y) / support(X)** |
|---|---|---|
| FA->FR | 0.28 | 0.59 |
| IS->FR | 0.17 | 0.25 |
| OA->FR | 0.06 | 0.25 |
| NE->FR | 0.08 | 0.375 |
| Friends and Others | 0 | - |
| Family and Internet Search | 0 | - |
| Family and Online Advertising | 0 | - |
| NE->FA | 0.19 | 0.875 |
| Family and Others | 0 | - |
| IS->OA | 0.11 | 0.17 |
| NE->IS | 0.06 | 0.25 |
| Internet Search and Others | 0 | - |
| OA->NE | 0.06 | 0.25 |
| Online Advertising and Others | 0 | - |

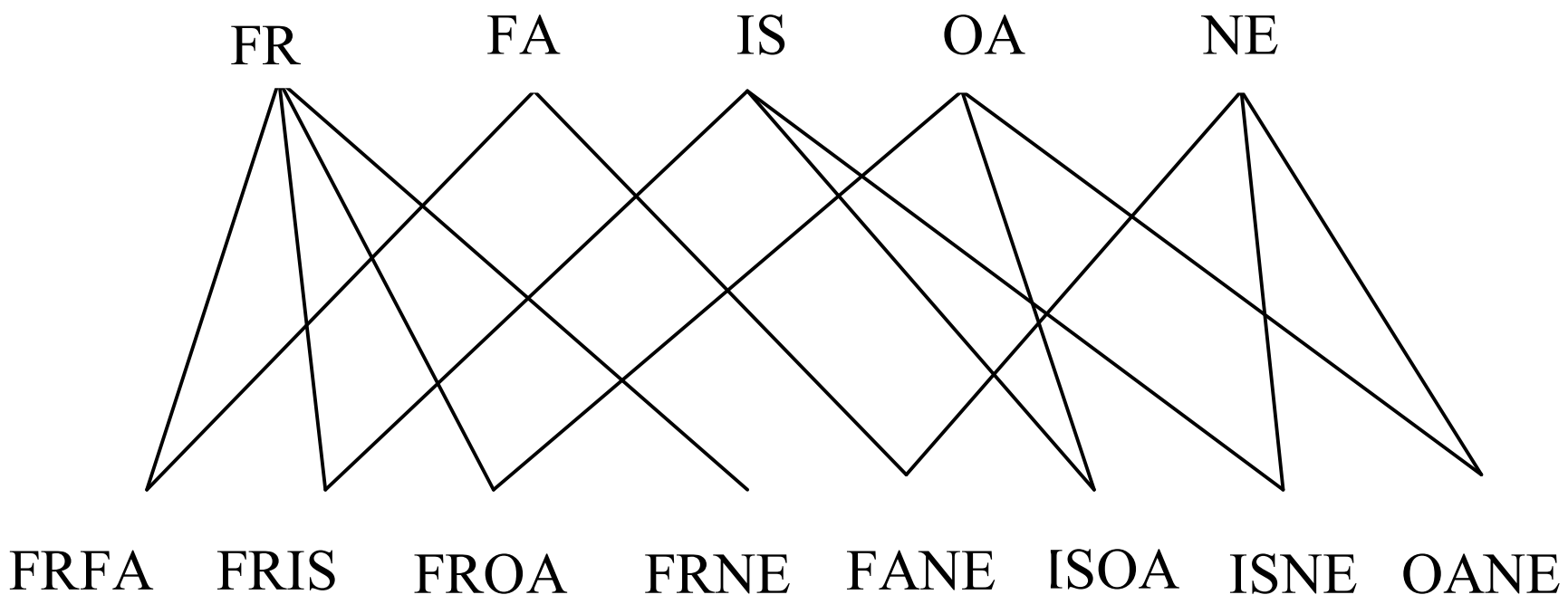

**Figure 3.21: Advertisement method's links**

# 3.2.3 Cosine Value Analysis

Cosine value occurs between 0 and 1.If its value is close to 1 then a good correlation occurs between X and Y. Cosine analysis formula defined as:
**Cosine(X->Y) = p(X,Y) / sqrt(p(X)*p(Y))**

**Table 3.13: Cosine analysis of different relations**

| **Relations** | **Cosine value=(p(X,Y)/sqrt(p(X)*p(Y)))** |
|---|---|
| FA->FR | 0.13 |
| IS->FR | 0.06 |
| OA->FR | 0.04 |
| NE->FR | 0.06 |
| **NE->FA** | **0.60** |
| IS->OA | 0.28 |
| NE->IS | 0.14 |
| OA->NE | 0.25 |

From table 3.12 result it can conclude that 28% students have answered with family and friends. Since 28% is largest support then it is a better advertisement method and its confidence value is 59% shows that if 59% of the time family occurs then friends also occurs in response. Figure 3.21 concluded that the all advertisement methods are linked to either friend advertisement or newspaper advertisement medium.

In figure 3.21 it is shown that friends and newspaper have five edges. In table 3.13, cosine values are calculated for all advertisement mediums and it concludes that **newspaper → family** has good relation to each other but others relations are not so close to each other.

# 3.2.4 Apriori algorithm implementation

**Table 3.14: Advertisement Methods data set**

| Relations | Code | Occurrences(Support count) |
|---|---|---|
| Friends | FR | 309 |
| Family | FA | 17 |
| Internet Search | IS | 24 |
| Online Advertising | OA | 8 |
| Newspaper | NE | 8 |
| Others | OT | 5 |
| Friends and Family | FRFA | 10 |
| Friends and Internet Search | FRIS | 6 |
| Friends and Online Advertising | FROA | 2 |
| Friends and Newspaper | FRNE | 3 |
| Family and Newspaper | FANE | 7 |
| Internet Search and Online Advertising | ISOA | 4 |
| Internet Search and Newspaper | ISNE | 2 |
| Online Advertising and Newspaper | OANE | 2 |

**Scan Database for count of each candidate item set**

**Step 1**: In this step the table 5 is scanned by algorithm and obtained information generates candidate item set $C_1$ (shown in table 6).
**Step 2:** In this step, compare $C_1$ candidate item set with minimum support count that generate frequent item set $L_1$

## *Apriori algorithm $C_1$ Join $L_1$ transformation*

**Table 3.15: Generated Candidate item set $C_1$**

| Code | Occurrences (Support count) |
|---|---|
| FR | 309 |
| FA | 17 |
| IS | 24 |
| OA | 8 |
| NE | 8 |
| OT | 5 |

**Minimum support count=5**

**= >Minimum support (5)**

**Table 3.16: Generated frequent item sets $L_1$**

| Code | Occurrences (Support count) |
|---|---|
| FR | 309 |
| FA | 17 |
| IS | 24 |
| OA | 8 |
| NE | 8 |
| OT | 5 |

**Step 3:** In step 3, $L_1$ Join $L_1$ provides $L_2$ frequent item set and candidate item set $C_2$ it is called joining process.

| Code | Occurrences (Support count) |
|---|---|
| FR | 309 |
| FA | 17 |
| IS | 24 |
| OA | 8 |
| NE | 8 |
| OT | 5 |

***$L_1$ Join $L_1$***

**Table 3.17: Generated Candidate set $C_2$**

| $L_1$*$L_1$ | $L_1$-1 | $L_1$-2 |
|---|---|---|
| FR,FA | FR | FA |
| FR,IS | FR | IS |
| FR,OA | FR | OA |
| FR,NE | FR | NE |
| FR,OT | FR | OT |
| FA,IS | FA | IS |
| FA,OA | FA | OA |
| FA,NE | FA | NE |
| FA,OT | FA | OT |
| IS,OA | IS | OA |
| IS,NE | IS | NE |
| IS,OT | IS | OT |
| OA,NE | OA | NE |
| OA,OT | OA | OT |
| NE,OT | NE | OT |

***Candidate item set $C_2$ generation by $L_1$ Join $L_1$ transformation***

**Step 4**: In this step the table 3.19 is scanned by algorithm and compare $C_2$ candidate item set with minimum support count that generate frequent item set $L_2$

**Table 3.18: Candidate set $C_2$ with Support Count**

| Code | Support Count |
|---|---|
| FR,FA | 10 |
| FR,IS | 6 |
| FR,OA | 2 |
| FR,NE | 3 |
| FR,OT | 0 |
| FA,IS | 0 |
| FA,OA | 0 |
| FA,NE | 7 |
| FA,OT | 0 |
| IS,OA | 4 |
| IS,NE | 2 |
| IS,OT | 0 |
| OA,NE | 2 |
| OA,OT | 0 |
| NE,OT | 0 |

**= > Minimum support (5)** →

**Table 3.19: Generated frequent item sets $L_2$**

| Code | Support count |
|---|---|
| FR,FA | 10 |
| FR,IS | 6 |
| FA,NE | 7 |

**Step 5:** In step 5, $L_2$ Join $L_2$ provides $L_3$ frequent item set and candidate item set $C_3$ (shown in table 11).

| Code | Support count |
|---|---|
| FR,FA | 10 |
| FR,IS | 6 |
| FA,NE | 7 |

→ ***$L_2$ Join $L_2$***

**Table 3.20: Generated Candidate set $C_3$**

| $L_2$*$L_2$ | $L_2$-1 | $L_2$-2 |
|---|---|---|
| FR,FA,IS | FR,FA | FR,IS |
| FR,FA,NE | FR,FA | FA,NE |
| FR,IS,FA,NE | FR,IS | FA,NE |

***Candidate item set $C_3$ generation by $L_2$ Join $L_2$ transformation***

**Step 6:**

- The subsets of (FR,FA,IS) are (FR,FA),(FR,IS)and(FA,IS).(FA,IS) is not a member of $L_2$. It is not a frequent item set. Therefore, remove (FR,FA,IS) from $C_3$.
- The subsets of (FR,FA,NE) are (FR,FA),(FA,NE)and(FR,NE).(FR,NE) is not a member of $L_2$. It is not a frequent item set. Therefore, remove (FR,FA,NE) from $C_3$.

- The subsets of (FR,IS,FA,NE) are
- (FR,IS),(FR,FA),(FR,NE),(IS,FA),(IS,NE)and(FA,NE) .(FR,NE),(IS,FA),(IS,NE) are not a members of $L_2$. It is not a frequent item set. Therefore, remove (FR, IS, FA, NE) from $C_3$.

- These item sets are pruned according to pruning property because its item sets are not frequent ,$C_3$ is null, so final frequent item set is $L_2$.If the minimum confidence is 50% then Confidence (FA->FR) and Confidence (NE->FA) is maximum so these are the frequent item sets (shown in table 3.22).

**Table 3.21: Frequent item set $L_2$**

| Code | Support Count | Confidence |
|---|---|---|
| FRFA | 10 | 0.59(59%) |
| FRIS | 6 | 0.25(25%) |
| FANE | 7 | 0.875(87%) |

**Table 3.22: Frequent item sets**

| Code | Confidence |
|---|---|
| FRFA | 0.59(59%) |
| FANE | 0.875(87%) |

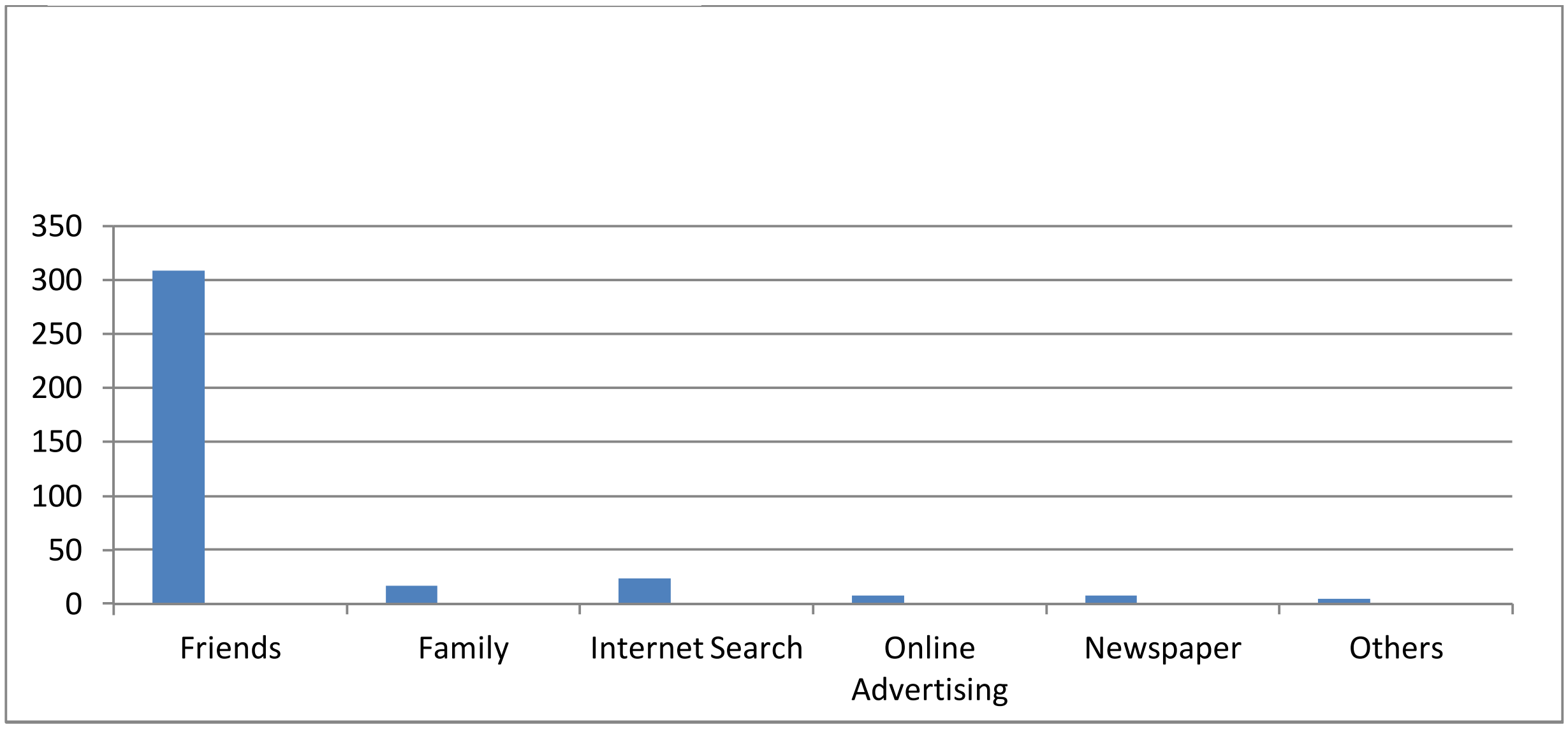

**Figure 3.22: Analysis chart that shows different advertisement methods(answered given in questionnaires)**

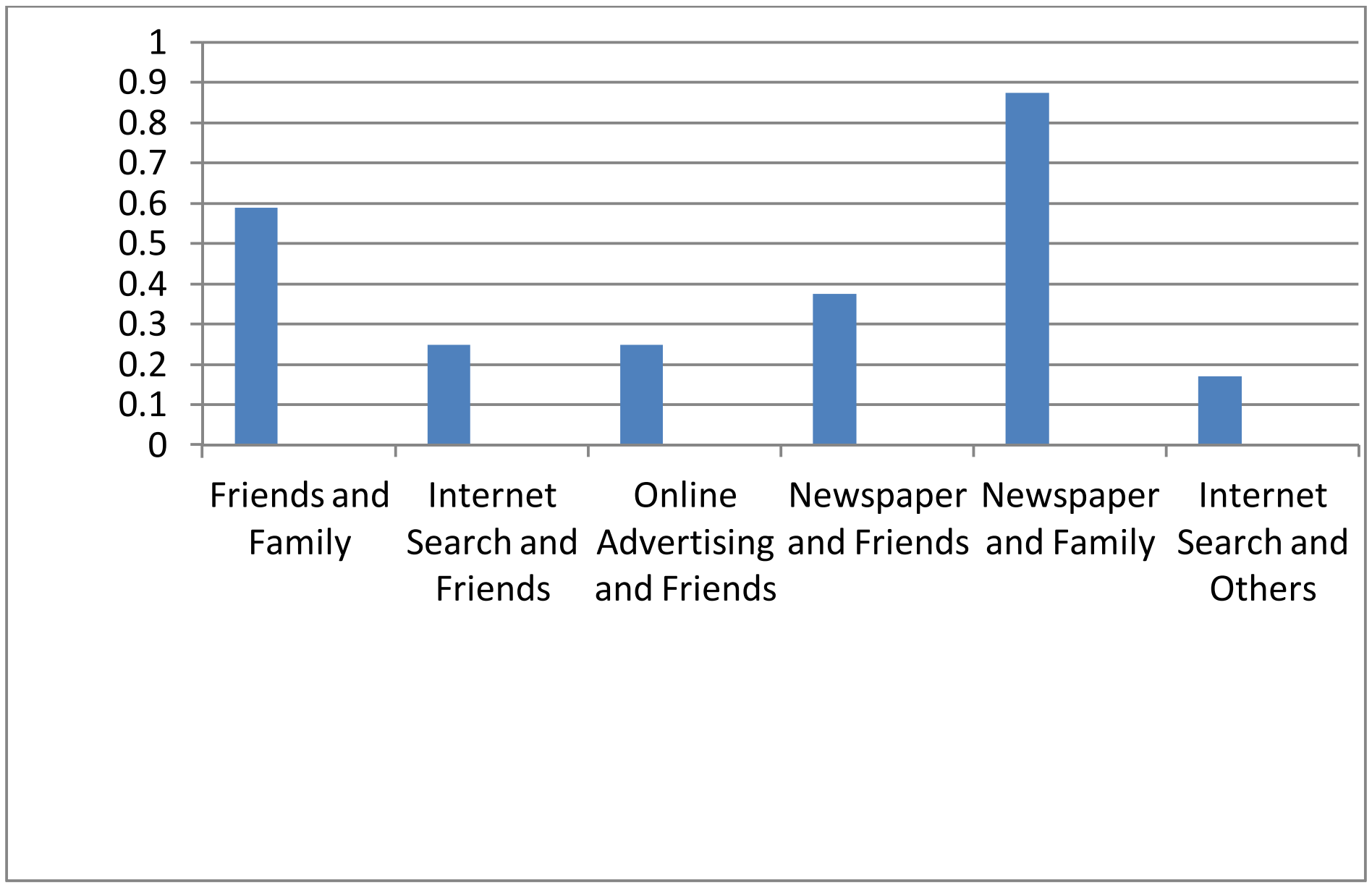

**Figure 3.23: Analysis chart that shows different advertisement methods (according to data mining analysis)**

# 3.2.5 Conclusion

This study looks for a better advertisement method. Fig.3.22 shows that most students have answered with friends (309 times) in the questionnaire but in this study data mining techniques are applied and this analysis concluded that the Family and Newspaper could be the best advertising method. Table 3.10 shows that most students have answered with friends (309 times) but with data mining technique's analysis it is concluded that the Friends and Family, Family and Newspaper could be the best advertisement methods because these advertisement methods are linked with all advertisement methods. After applying apriori algorithm on the data set, Table 3.22 shows frequent item sets (FRFA,FANE) that concluded that Friends and Family, Family and Newspaper is the best advertisement methods. This study is helpful for educational institutes to making a good advertisement strategy that attracts student effectively.

## 3.3 Classification of Student's data Using Data Mining Techniques for Training & Placement Purpose.

Data Mining is new approach for technical education. Technical institute like engineering & other can use data mining techniques for analysis of different performances in student's qualifications. In our work, we collected enrolled student's data from engineering institute that have different information like 10th, 12th, B.tech passing percentage and then apply decision tree method for classifying students academics performance for Training & placement department can be identify the final grade of student for placement purpose. In future this study will be help to develop new approaches of data mining techniques in technical education.

### 3.3.1 Data Used

**Table 3.23: Student Attribute Description**

| Attribute | Possible Values |
|---|---|
| Branch | CS/IT |
| $10^{th}$ Per | { First > 60%,Second > 45 & < 60 %,Third > 35 & < 45 % } |
| $12^{th}$ Per | { First > 60%,Second > 45 & < 60 %,Third > 35 & < 45 % } |
| B.Tech Per | { First > 60%,Second > 45 & < 60 %,Third > 35 & < 45 % } |
| Final Grade | { Excellent, Good, Average } |

### 3.3.2 Classification Rules

IF 10th % ="First" AND 12th % ="First" AND B.Tech % = "First" THEN Final_Grade = "Excellent"
IF 10th % ="Second" AND 12th % ="First" AND B.Tech % = "First" THEN Final_Grade = "Good"
IF 10th % ="Third" AND 12th % ="First" AND B.Tech % ="First" THEN Final_Grade = "Average"
IF 10th % ="First" AND 12th % ="Second" AND B.Tech % ="First" THEN Final_Grade = "Good"
IF 10th % ="Second" AND 12th % ="Second"AND B.Tech %= "First" THEN Final_Grade = "Average"
IF 10th % ="Third" AND 12th % ="Second" AND B.Tech %= "First" THEN Final_Grade = "Average"
IF 10th % ="First" AND 12th % ="First" AND B.Tech % ="Second" THEN Final_Grade = "Average"
IF 10th % ="Second" AND 12th % ="First" AND B.Tech % ="Second" THEN Final_Grade = "Average"

IF 10th % ="Third" AND 12th % ="First" AND B.Tech % ="Second" THEN Final_Grade = "Average"
IF 10th % ="First" AND 12th % ="Second" AND B.Tech % ="Second" THEN Final_Grade = "Average"
IF 10th % ="Second" AND 12th % ="Second"AND B.Tech %= "Second" THEN Final_Grade = "Average"

## 3.3.3 Conclusion & Future Work

In this work we make use of data mining process in a student's database using classification data mining techniques (decision tree method). The information generated after the analysis of data mining techniques on student's data base is helpful for executives for training & placement department of engineering colleges. This work classifies the categories of student's performance in their academic qualifications. For future work, this study will be helpful for institutions and industries. We can be generating the information after implementing the others data mining techniques like clustering, Predication and Association rules etc on different eligibility criteria of industry recruitment for students.

# CHAPTER 4